\documentclass[10pt,twocolumn,letterpaper]{article}

\usepackage[pagenumbers]{cvpr} %

\usepackage[T1]{fontenc}
\usepackage{float}
\usepackage{xspace}
\usepackage{algorithm}
\usepackage{algpseudocode}
\usepackage{multirow}
\newcommand{\alg}{\textsc{Progressor}\xspace}

\definecolor{myorange}{rgb}{0.937,0.525,0.211}
\definecolor{myblue}{rgb}{0.231,0.459,0.686}

\definecolor{mygreen}{rgb}{0.4,0.8,0}
\definecolor{myred}{rgb}{1,0.2,0.2}

\newcommand{\xline}[1]{$\rule[0.5ex]{#1}{0.4pt}$}

\definecolor{cvprblue}{rgb}{0.21,0.49,0.74}
\usepackage[pagebackref,breaklinks,colorlinks,allcolors=cvprblue]{hyperref}

\def\paperID{10544} %
\def\confName{CVPR}
\def\confYear{2025}

\usepackage{xspace}
\definecolor{celestialblue}{rgb}{0.29, 0.59, 0.82}
\newif\ifshowcomments

\ifshowcomments
    \newcommand{\matt}[1]{\textcolor{BrickRed}{\textbf MW: #1}\xspace}
    \newcommand{\xiao}[1]{\textcolor{celestialblue}{\textbf XZ: #1}\xspace}
    \newcommand{\mmaire}[1]{\textcolor{purple}{\textbf MM: #1}\xspace}
    \newcommand{\tc}[1]{\textcolor{celestialblue}{\textbf TC: #1}\xspace}
    \newcommand{\kw}[1]{\textcolor{orange}{\textbf KW: #1}\xspace}
\else
    \newcommand{\matt}[1]{}
    \newcommand{\xiao}[1]{}
    \newcommand{\mmaire}[1]{}
    \newcommand{\tc}[1]{}
    \newcommand{\kw}[1]{}
\fi
\title{\alg: A Perceptually Guided Reward Estimator\\with Self-Supervised Online Refinement}

\author{Tewodros W. Ayalew$^{1}$~~~Xiao Zhang$^{*,1}$~~~Kevin Yuanbo Wu$^{*,1}$~~~Tianchong Jiang$^{2}$\\Michael Maire$^{1}$~~~Matthew R. Walter$^{2}$\\
\\[-7pt]
$^1$University of Chicago\quad
$^2$Toyota Technological Institute at Chicago\\
\\[-7pt]
\fontsize{11pt}{\baselineskip}\selectfont{\tt\href{https://ripl.github.io/progressor}{\textbf{https://ripl.github.io/progressor}}}
}

\usepackage{amsmath,amsfonts,bm}

\def\eqref#1{equation~\ref{#1}}

\def\1{\bm{1}}

\def\vo{{\bm{o}}}

\def\vr{{\bm{r}}}

\def\mE{{\bm{E}}}

\DeclareMathAlphabet{\mathsfit}{\encodingdefault}{\sfdefault}{m}{sl}
\SetMathAlphabet{\mathsfit}{bold}{\encodingdefault}{\sfdefault}{bx}{n}

\begin{document}
\maketitle
\def\thefootnote{*}\footnotetext{Equal contribution.} \def\thefootnote{\arabic{footnote}}
\begin{abstract}
We present \alg, a novel framework that learns a task-agnostic reward function from videos, enabling policy training through goal-conditioned reinforcement learning (RL) without manual supervision. Underlying this reward is an estimate of the distribution over task progress as a function of the current, initial, and goal observations that is learned in a self-supervised fashion. Crucially, \alg refines rewards adversarially during online RL training by pushing back predictions for out-of-distribution observations, to mitigate  distribution shift inherent in non-expert observations. Utilizing this progress prediction as a dense reward together with an adversarial push-back, we show that \alg enables robots to learn complex behaviors without any external supervision. Pretrained on large-scale egocentric human video from EPIC-KITCHENS, \alg requires no fine-tuning on in-domain task-specific data for generalization to real-robot offline RL under noisy demonstrations, outperforming contemporary methods that provide dense visual reward for robotic learning. Our findings highlight the potential of \alg for scalable robotic applications where direct action labels and task-specific rewards are not readily available.
\end{abstract}

\section{Introduction}\label{sec:intro}
\begin{figure}[t]
    \centering
    \captionsetup{type=figure}
    \includegraphics[width=.5\textwidth]{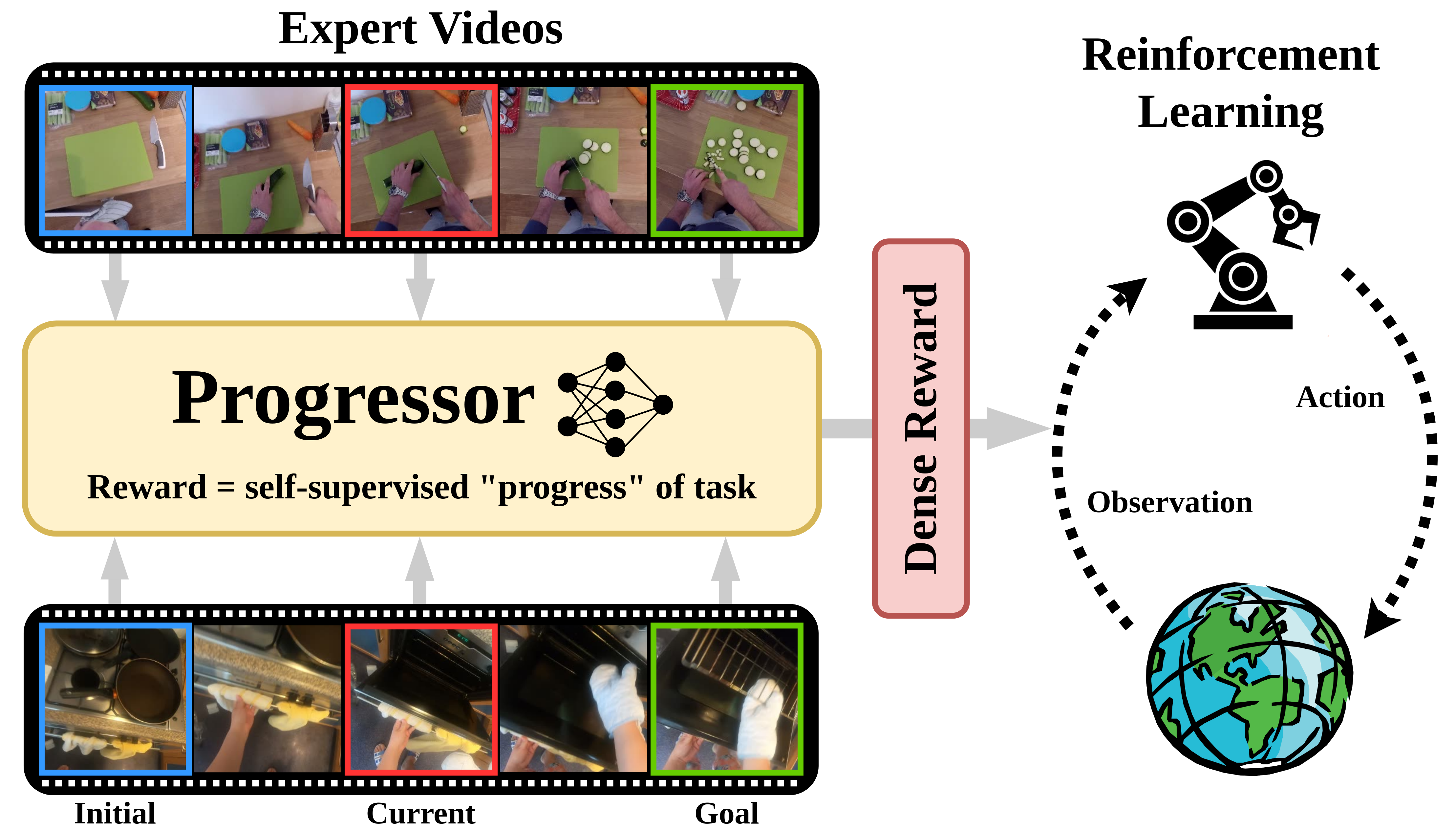}
    \captionof{figure}{
    Trained in a self-supervised manner on expert videos, \alg predicts an agent's progress toward task completion, providing a reward signal for reinforcement learning. During online reinforcement learning, we employ an adversarial technique to refine this reward estimate, addressing the distribution shift between expert data and non-expert online rollouts.}
    \label{fig:cover-figure}
\end{figure}%

Practical applications of %
reinforcement learning (RL) require that a domain expert design %
complex reward models that encourage desired behavior while simultaneously penalizing unwanted behavior~\citep{RussellNorvig1995, Singh2010Rewards, SuttonBarto2018}. However, manually constructing dense rewards that enable effective learning is difficult and can yield undesired behaviors~\citep{HadfieldMenell2017Inverse, Booth2023Perils}. Sparse rewards require far less effort, however the lack of supervision typically results in a significant loss of sample efficiency. These issues are exacerbated for long-horizon tasks, where credit assignment is particularly difficult. %

A promising alternative is to learn effective rewards from unlabeled (i.e., action-free) videos of task demonstrations~\citep{nair2022r3muniversalvisualrepresentation, ma2023vipuniversalvisualreward, alakuijala2023learningrewardfunctionsrobotic, yang2024rank2reward, huang2023diffusion, escontrela2024video}. Large-scale video data is readily available on the internet, providing a effective alternative to costly, intensive task-specific data collection. An effective way to incorporate diverse video data into a \emph{single} model without learning a new policy for each task is to condition the model on a goal image that visually specifies the desired environment changes upon task completion~\citep{eysenbach2023contrastivelearninggoalconditionedreinforcement, ma2023vipuniversalvisualreward}.

To that end, we propose \alg, a \textit{self-supervised} framework that learns a \emph{task-agnostic} reward function from videos for \textit{goal-conditioned reinforcement learning} that only requires image observations, without the need for corresponding action labels. Following \citet{yang2024rank2reward}, we assume that expert demonstrations make monotonic progress towards a goal and employ a self-supervised temporal learning approach that utilizes a proxy objective aimed at estimating task completion progress. \alg predicts the distribution of progress from the current observation (i.e., the current image) relative to the initial and goal observations. This estimation provides dense supervision of an agent's progression towards reaching the goal, and thus performing the task, guiding exploration in alignment with expert execution trajectories. However, during RL exploration, agents will often encounter states that were not visited as part of the expert trajectories. \alg accounts for this distribution shift via an online adversarial refinement that pushes back pushing back predictions for out-of-distribution observations. 

We evaluate the effectiveness of \alg for robot manipulation tasks in both simulation and the real world. These results demonstrate the the benefits that \alg's self-supervised reward has on policy learning, particularly in settings where existing methods otherwise require intricate reward design.  
The key contributions of this paper are as follows:

\begin{itemize}
    \item We present \alg, a self-supervised reward model trained on unlabeled videos that guides a reinforcement learning agent by providing dense rewards based on predicted progress toward a goal.%
    \item \alg achieves state-of-the-art performance on six diverse tasks from the Meta-World benchmark~\citep{yu2020meta} without the need for environment reward.
    \item Pretrained on large-scale egocentric human video from EPIC-KITCHENS~\citep{Damen2018EPICKITCHENS}, \alg enables few-shot real-robot offline, \textit{even when half of the demonstrations are unsuccessful}.
\end{itemize}

\section{Related Work}\label{sec:related-work}

Learning from expert videos has gained traction due to the potential to scale the acquisition of complex skills using the vast amount of video content readily available. %

\textbf{Inverse RL} Inverse reinforcement learning (IRL) is a framework for learning reward functions from expert demonstrations as an alternative to specifying them manually. In contrast to traditional RL, where the reward function is predefined and the goal is to learn an optimal policy, IRL focuses on deducing the underlying reward structure that results in the expert's behavior~\citep{wulfmeier2016maximumentropydeepinverse, 10.1145/1015330.1015430, 10.5555/1620270.1620297}. Recent IRL methods leverage adversarial techniques to learn reward functions~\cite{fu2018learningrobustrewardsadversarial, li2017infogailinterpretableimitationlearning, finn2016guidedcostlearningdeep}, such as in Generative Adversarial Imitation Learning (GAIL)~\cite{ho2016generativeadversarialimitationlearning}. A key limitation of typical approaches to IRL is that they require action-labeled demonstrations, which are only available in small quantities. Another is the ambiguity that results from having multiple rewards that can explain the observed behavior~\citep{metelli2023theoreticalunderstandinginversereinforcement}, potentially leading to ill-shaped reward functions. In contrast, our notion of progress enables the learning of well-shaped dense rewards. 

\textbf{Imitation learning from video} 
Several methods~\citep{zhang2022learning, baker2022video} learn from videos by inferring the inverse dynamics model, leveraging a small amount of annotated data. However, these methods depend on the presence and quality of an albeit limited action labels. 
Other methods have used existing hand pose estimation as intermediate annotation followed by standard imitation learning~\citep{bahl2022human, qin2022dexmv}. However, these methods rely on the accuracy of the pose estimators and camera calibration.

\textbf{Learning reward from video}
Recent methods learn visual representations useful for RL from video data using self-supervised temporal contrastive techniques~\citep{ma2023vipuniversalvisualreward, nair2022r3muniversalvisualrepresentation, sermanet2018timecontrastivenetworksselfsupervisedlearning, eysenbach2023contrastivelearninggoalconditionedreinforcement, yang2021representationmattersofflinepretraining}. Time-contrastive objectives treat observations close in time as positive pairs and those that are temporally distant as negative pairs. They then learn embeddings that encourage small distances between positive pairs and large distances between negative pairs. The learned embeddings can then be used to define reward functions, i.e., as the distance from the goal image in embedding space. Time-contrastive objectives can be sensitive to frame rate, and extracted rewards are assumed to be symmetric~\citep{ma2023vipuniversalvisualreward}. In comparison, our approach is agnostic to the sampling rate and can learn asymmetric dynamics depending on the specified initial and goal images.

Additionally, generative approaches have been used for reward learning. VIPER~\citep{escontrela2024video} estimates the log-likelihood of an observation as a reward function by training on an expert dataset using a VideoGPT-based autoregressive model. Similarly, \citet{huang2023diffusion} employ a diffusion-based approach to learn rewards by leveraging conditional entropy. However, these methods necessitate costly generative processes to estimate rewards.

Particularly relevant, Rank2Reward~\citep{yang2024rank2reward} demonstrates that learning the ranking of visual observations from a demonstration helps to infer well-shaped reward functions. By integrating ranking techniques with the classification of expert demonstratin data vs.\ non-expert data obtained through on-policy data collection, Rank2Reward guides robot learning from demonstration. A key limitation of the method is that it necessitates training distinct models for each task due to the task-dependent nature of ranking frames. In contrast, our approach employs a \emph{single} reward model for all tasks. Additionally, we utilize an adversarial training approach that tackles the domain shift observed during online RL, rather than incorporating a separate classifier as used in Rank2Reward.

\section{Preliminaries}\label{sec:preliminaries}

We formulate our method in the context of a finite-horizon Markov decision process (MDP) $\mathcal{M}$ defined by the tuple ($\mathcal{O}, \mathcal{A}, \mathcal{P}, \mathcal{R}, \mathcal{\gamma}, \mathcal{\rho}_{0}$), which represents the observation space, action space, reward function, transition probability function, discount factor, initial state distribution, respectively. Typically, the agent starts from an initial state with observation $\vo_0 \sim \mathcal{O}$ and at each step performs an action $a \sim \mathcal{A}$. %
The fundamental objective in RL is to learn optimal policy $\pi^*$ that maximizes the expected discounted reward, i.e., expressed as $\pi^* = \operatorname*{argmax}_{\pi}\mathbb{E}_{\pi_{\theta}}[{\sum^T_{t=1}(\gamma^{t-1} \mathcal{R}(\vo_t))}]$. 

In this paper, we operate under the assumption that the environment is fully observable, wherein observations \mbox{$\vo_k \in \mathbb{R}^{H \times W \times 3}$} are high-dimensional image frames. Additionally, we assume access to a dataset $\mathcal{D}_e$ that consists of expert demonstrations. These demonstrations are comprised of sets of trajectories ${\{\mathcal{\tau}_k\}^{N}_{k=1}}$, each containing sequences of observations, $\mathcal{\tau}_k = \{\vo_0^{\mathcal{\tau}_k}, \vo_1^{\mathcal{\tau}_k}, ..., \vo_n^{\mathcal{\tau}_k}\}$. 

However, we assume that neither the reward function $\mathcal{R}$ nor the demonstrated actions $a \sim \mathcal{A}$ are known. This setup mirrors the real-world robot task-learning scenario in which we have a plethora of videos of human-performed tasks, without knowledge of the actions being performed or the rewards being received. The primary objective of this paper is to construct a model capable of predicting a well-structured dense reward from a dataset of expert demonstrations, without access to demonstrated actions or rewards. This model can subsequently serve as a reward estimating mechanism when training an agent using online RL in the absence of
any environment rewards.

\section{Method}\label{sec:method}
\begin{figure*}[!t]
    \centering
    \includegraphics[width=0.85\linewidth]{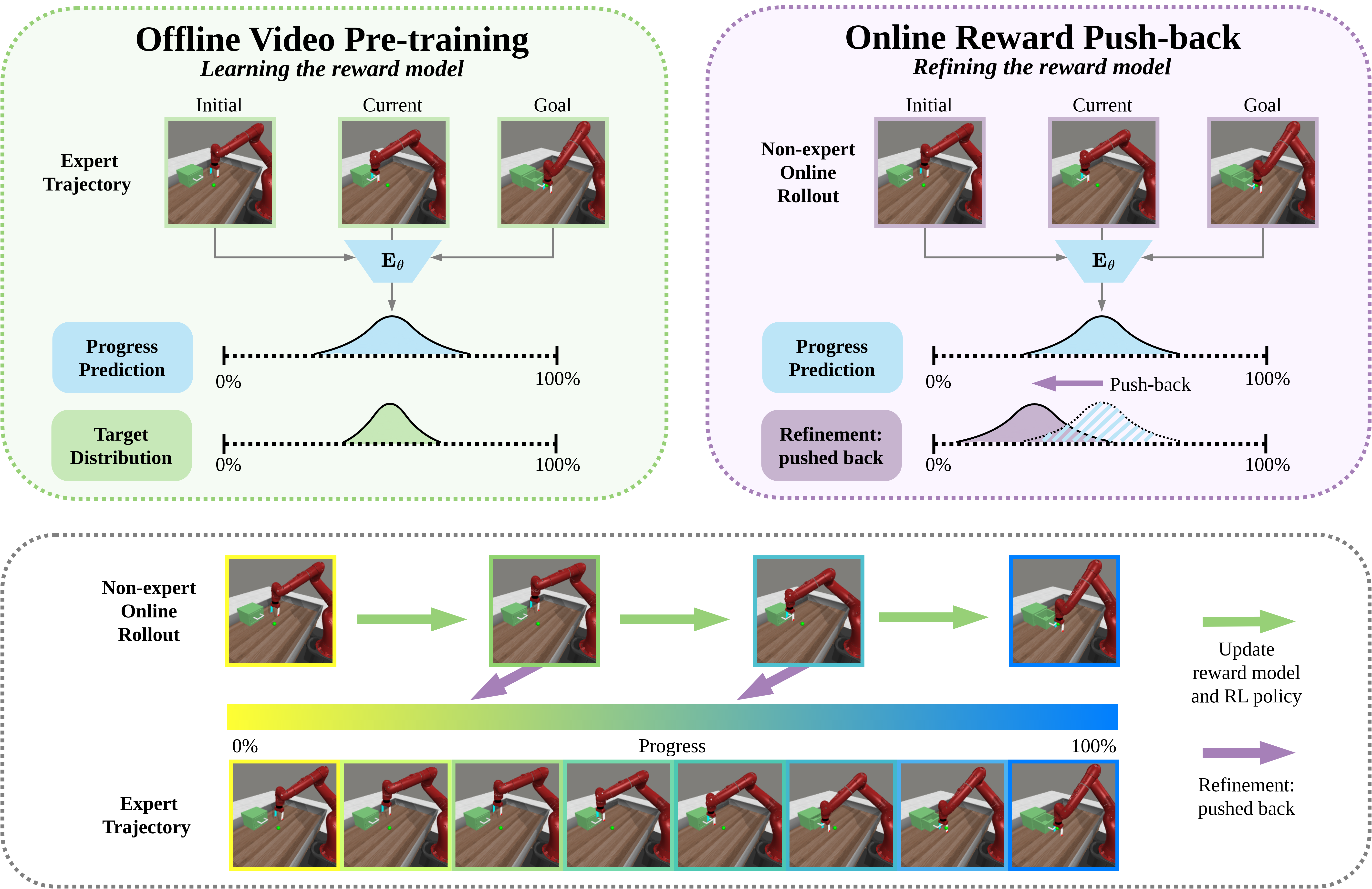}
    \caption{\textbf{Top Left:} Initial phase of reward model pretraining on expert data, where the model learns to predict the parameters of a  Gaussian distribution centered on normalized progress, reflecting expected progress as demonstrated by experts. \textbf{Top Right:} In online reinforcement learning (RL) training, an adversarial online refinement (i.e., push-back) is applied to counteract non-expert predictions made by the reward model, effectively distinguishing expert from non-expert progress. \textbf{Bottom:} During online RL, the reward model is updated on expert and non-expert data.
    }
    \label{fig:progressor-method}
\end{figure*}
We propose to learn a unified reward model via an encoder $\mE_{\theta}(\vo_i, \vo_j, \vo_g)$ that estimates the relative progress of an observation $\vo_j$ with respect to an initial observation $\vo_i$ and a goal observation $\vo_g$, all of which are purely pixel-based. Figure~\ref{fig:progressor-method} illustrates \alg's framework for task progress estimation. Here, the objective is to learn the temporal ordering of observations that lead to the correct execution of a task. By learning this progress estimation, we can create a reward model that incentivizes progress toward the goal. 
 Similar to \citet{yang2024rank2reward}, our method relies on the assumption that the values of states in optimal policies increase monotonically towards task completion.

\subsection{Learning the Self-Supervised Reward Model}
\label{sec: learning_reward_model}
We measure an observation $\vo_j$'s progress towards a  goal in terms of its relative position with respect to the initial observation $\vo_i$ and the goal observation $\vo_g$. In order to learn this progress estimate, we first represent the relative position of an observation $\vo_j$ within an expert trajectory $\tau_k$ as 
\begin{equation}
    \label{eqn:distance}
    \delta(\vo_i^{\tau_k},\vo_j^{\tau_k}, \vo_g^{\tau_k}) = \frac{ \left|j - i\right|}{ \left|g - i\right|}.
\end{equation}
Here, $\delta(\vo_i^{\tau_k}, \vo_j^{\tau_k}, \vo_g^{\tau_k})$ calculates the ratio of the relative frame position differences between $\vo_j$ and $\vo_i$ and between $\vo_g$ and $\vo_i$ within a given expert trajectory $\tau_k$. This progress label is always such that $\delta(\vo_i^{\tau_k}, \vo_j^{\tau_k}, \vo_g^{\tau_k}) \in [0, 1]$. 

Predicting progress based on real-world video is challenging due to factors like camera motion and repetitive sub-trajectories. 
We address this by formulating the problem as one of estimating a Gaussian distribution over progress, allowing the model to estimate uncertainty as variance. 
With this approach, we approximate the ground-truth distribution for a triplet of frames $(\vo_i^{\tau_k}, \vo_j^{\tau_k}, \vo_g^{\tau_k})$ within a video $\tau_k$ as a normal distribution with mean $\mu_{\tau_k} = \delta(\vo_i^{\tau_k}, \vo_j^{\tau_k}, \vo_g^{\tau_k})$ and standard deviation $\sigma_{\tau_k} = \max(\frac{1}{g - i}, \epsilon)$:
\begin{equation}
    \label{eqn:progress-dist}
    p_{\text{target}}(\vo_i^{\tau_k}, \vo_j^{\tau_k}, \vo_g^{\tau_k})= \mathcal{N}\left(\mu_{\tau_k}, \sigma_{\tau_k}^2\right)
\end{equation}
Our $\epsilon$ upper-bound of $\sigma_{\tau_k}$ downweights the penalty for triplets in which there are many frames between the initial and goal images, 
which we empirically find to improve robustness during training. %
Therefore, the predicted distribution of this form provides a distribution of the progress estimate.
Sampling from this predicted distribution yields values that reflect the observed progress.

We optimize our reward model $\vr_{\theta}$ to predict the distribution of the progress on expert trajectory. 
We use a shared visual encoder to compute the per-frame representation, followed by several MLPs to produce the final estimation:
\begin{equation} \label{eqn:e_output}
    \mE_{\theta}(\vo_i, \vo_j, \vo_g)= \mathcal{N}\left(\mu, \sigma^2\right)
\end{equation}
Following \citet{kingma2013auto}, we estimate $\log\sigma^2$ and then compute $\sigma$ as $\sigma = \exp\left(0.5\log\sigma^2\right)$.
We learn the parameters of the network by optimizing the Kullback-Leibler divergence loss between the ground-truth and the predicted distributions:
\begin{equation} \label{eqn:objective-function}
    \!\!\mathcal{L}_{\textrm{expert}} \!\!=\!\! D_\textrm{KL} \!\!\left(p_{\text{target}}\left(\vo_i^{\tau_k}, \vo_j^{\tau_k}, \vo_g^{\tau_k}\right) \!\Vert \mE_{\theta}\left(\vo_i^{\tau_k}, \vo_j^{\tau_k}, \vo_g^{\tau_k}\right)\!\right)\!
\end{equation}
During the training of our reward model $\vr_{\theta}$, we randomly select and rank three frames from a video sequence of expert trajectories as a training triplet. Additionally, since these triplets can be randomly drawn from any trajectory within the dataset of different robotics tasks, a single reward model suffices to handle a variety of tasks.

\begin{algorithm}[!ht]
  \caption{\alg in Online RL}\label{alg:progressor}
  \begin{algorithmic}[1]
    \Require Expert demonstration data $\mathcal{D}_e = \{\tau_k\}_{k=1}^N$
    \State Initialize policy $\pi$, empty replay buffer $\mathcal{D}_{RB}$
    \State Initialize \alg as ${\mathbf{r}}_{\theta}$.
    \State \textcolor{lightgray}{// Pretraining \alg}
    \For{step $n$ in $\{1, \dots, N_{\text{pretrain}}\}$}
      \State Sample triplets $(\vo_i^{\tau_k}, \vo_j^{\tau_k}, \vo_g^{\tau_k})$ from $\tau_k \in \mathcal{D}_e$
      \State Learn $\vr_{\theta}$ with objective Eqn.~\ref{eqn:objective-function}
    \EndFor
    \State \textcolor{lightgray}{// Policy optimization and reward model fine-tuning}
    \For{step $n$ in $\{1, \dots, N\}$}
      \State With $\pi$, collect transitions $\{\tau_l\}_{l=1}^M$ and store in $\mathcal{D}_{RB}$
      \If{$n \ \% \ reward\_update\_frequency == 0$}
        \State Update $\vr_{\theta}$ with Eqn.~\ref{eqn:objective-function} using samples from $\mathcal{D}_{e}$
        \State Update $\vr_{\theta}$ with Eqn.~\ref{eqn:push-back} using samples from $\mathcal{D}_{RB}$
      \EndIf
      \State Sample batch of transitions $s_\pi$ from $\mathcal{D}_{RB}$
      \State Update $\pi$ to maximize returns using Eqn.~\ref{eqn:reward}
    \EndFor
  \end{algorithmic}
\end{algorithm}

\textbf{Using the Reward Model in Online RL:} Due to the monotonic nature of task progress in our problem formulation, the estimates derived from the trained model can serve as dense rewards for training, replacing the unknown true reward. We create the reward model by defining  a function derived from the model's predicted outputs given a sample of frame triplet ($\vo_i, \vo_j, \vo_g$) of trajectory as:
\begin{equation} \label{eqn:reward}
    \vr_{\theta}(\vo_i, \vo_j, \vo_g) = \mu - \alpha \mathcal{H}(\mathcal{N}(\mu, \sigma^2)),
\end{equation}
where $\vr_{\theta}$ denotes \alg's reward estimate and $\mathcal{H}$ is the entropy. The first term $\mu$ in Eqn.~\ref{eqn:reward} uses the mean prediction from the progress estimation model to determine the position of the current observation relative to the goal. 
The second term penalizes %
observations that yield high variance. Here, $\alpha$ is hyperparameter, which we set to $\alpha = 0.4$ in all our experiments. %
See S8
for an ablation of $\alpha$.

\subsection{Adversarial Online Refinement via Push-Back}

Our reward model $\vr_{\theta}$ can be used to train a policy that maximizes the expected sum of discounted rewards as:%
\begin{align}
    \pi^* = \underset{\pi}{\textrm{arg max}} \; \mathbb{E}_{\pi_{\theta}}\left[{\sum^T_{t=1}(\gamma^{t-1} \vr_{\theta} (\vo_i^{\tau'}, \vo_t^{\tau'}, \vo_g^{\tau'})}\right]
\end{align}
The reward encourages the agent to take actions that align its observed trajectories $\tau'$ 
with those of the expert demonstrations $\tau$, thereby promoting task completion. However, the all-but-random actions typical of learned policies early in training result in out-of-distribution observations relative to the expert trajectories. As a result, optimizing the policy using a reward $\vr_{\theta}$ that is only trained on an expert trajectory $\tau$ can result in unintended behaviors, since $\vr_{\theta}$ does not reflect the actual progress of $\tau'$. %

To tackle this distribution shift, we implement an adversarial online refinement strategy~\cite{goodfellow2014generative,ganin2016domain}, which we refer to as ``push-back'', that enables the reward model $\vr_{\theta}$ to differentiate between in- and out-of-distribution observations $\tau$ and $\tau'$, providing better estimation. Specifically, for a frame triplet $(\vo_i^{\tau_k'}, \vo_j^{\tau_k'}, \vo_g^{\tau_k'})$ sampled from $\tau_k'$
and the estimated progress $\mu_{\tau_k'}$ from $\mE_{\theta}$ (\ref{eqn:e_output}), 
we update $\mE_{\theta}$ so that it learns to \emph{push-back} the current estimation as $\beta\mu_{\tau_k'}$ with $\beta \in [0,1]$ as the decay factor. In our experiments, we use $\beta = 0.9$. We formulate our online refinement objectives as:
\begin{align}
\label{eqn:push-back}
    p_{\textrm{push-back}} &= \textrm{sg}\left(\mathcal{N}\left(\beta\mu_{\tau_k'}, \frac{1}{(g - i)^2}\right)\right) \nonumber\\
    \mathcal{L}_{\textrm{push-back}} &= D_\textrm{KL} \left(p_{\textrm{push-back}} \Vert  \mE_{\theta}\left(\vo_i^{\tau_k'}, \vo_j^{\tau_k'}, \vo_g^{\tau_k'}\right)\right),
\end{align} 
where $\textrm{sg}(\cdot)$ denotes the stop gradient operator. With this approach, we differentiate between the two sources of trajectories by treating observations from the (inferior) online roll-out trajectories $\tau'$ as corresponding to lower progress compared to expert trajectories $\tau$. %
In this case, optimizing for $\pi^*$ would encourage the agent to consistently improve its actions until the policy trajectories are well aligned with the distribution of expert trajectories $\tau_k$ .

During online training, we fine-tune $\mE_{\theta}$ (\ref{eqn:e_output}) using hybrid objectives: we optimize $\mathcal{L}_{\textrm{push-back}}$ based on $\tau_k'$ sampled from the most recent buffers and at the same time, we optimize $\mathcal{L}_{\textrm{expert}}$ using samples from $\tau_k$, preventing $\mE_{\theta}$ from being biased by $\tau_k'$.

\section{Experimental Evaluation}\label{sec:experiments-and-results} \begin{figure}[t]
    \centering
    \captionsetup{type=figure}
    \includegraphics[width=.5\textwidth]{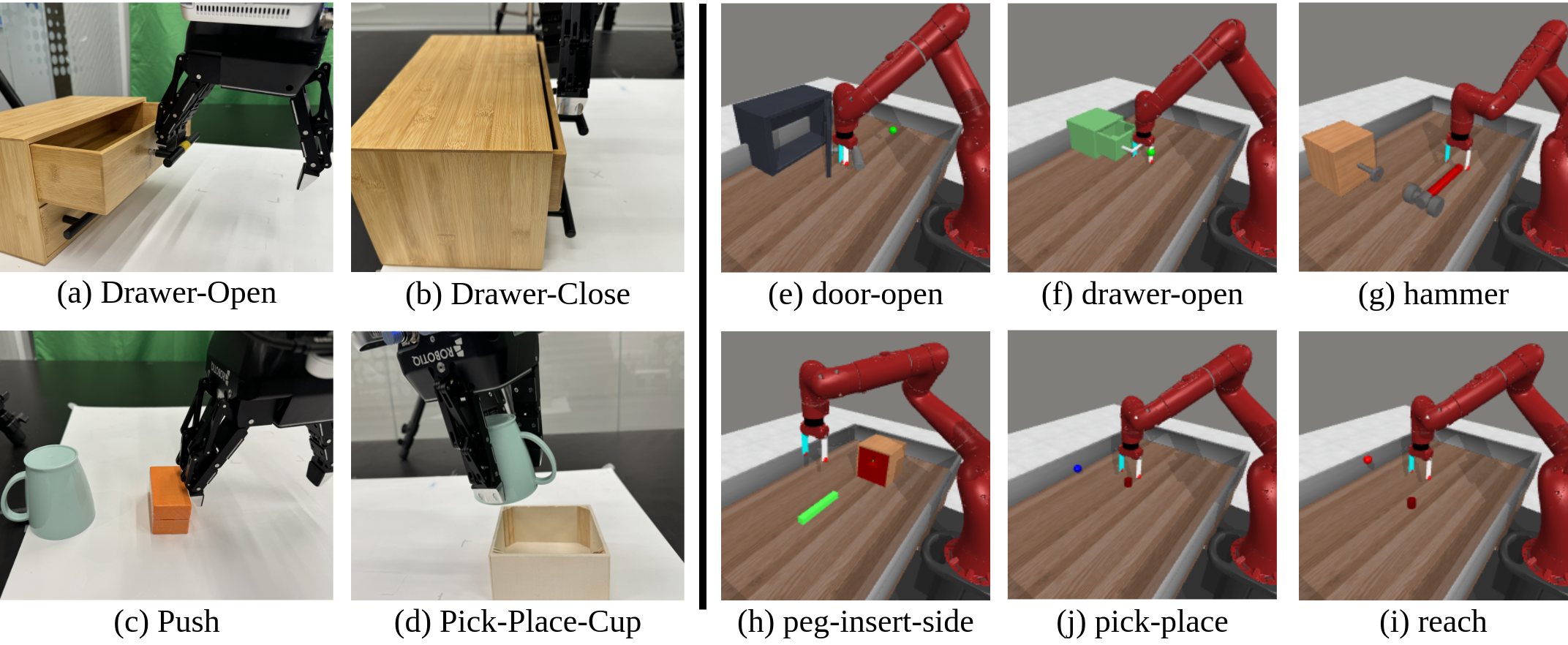}
    \captionof{figure}{Visualization of the robotic tasks: (a-d) Real world environments with a UR5 arm. (e-j) Simulation environments for evaluation using the Meta-World~\citep{yu2020meta} benchmark.}
    \label{fig:all-tasks}
\end{figure}%

We evaluate the effectiveness with which \alg learns reward functions from visual demonstrations that enable robots to perform various manipulation tasks in simulation as well as the real world. The results demonstrate noticeable improvements over the previous state-of-the-art. %

\subsection{Simulated Experiments} \begin{figure*}[!ht]
\centering
\begin{subfigure}{0.9\linewidth}
    \centering
    \includegraphics[width=1.0\linewidth]{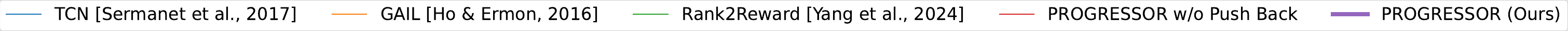}
\end{subfigure}
\begin{subfigure}{0.48\linewidth}
    \centering
    \includegraphics[width=1.0\linewidth]{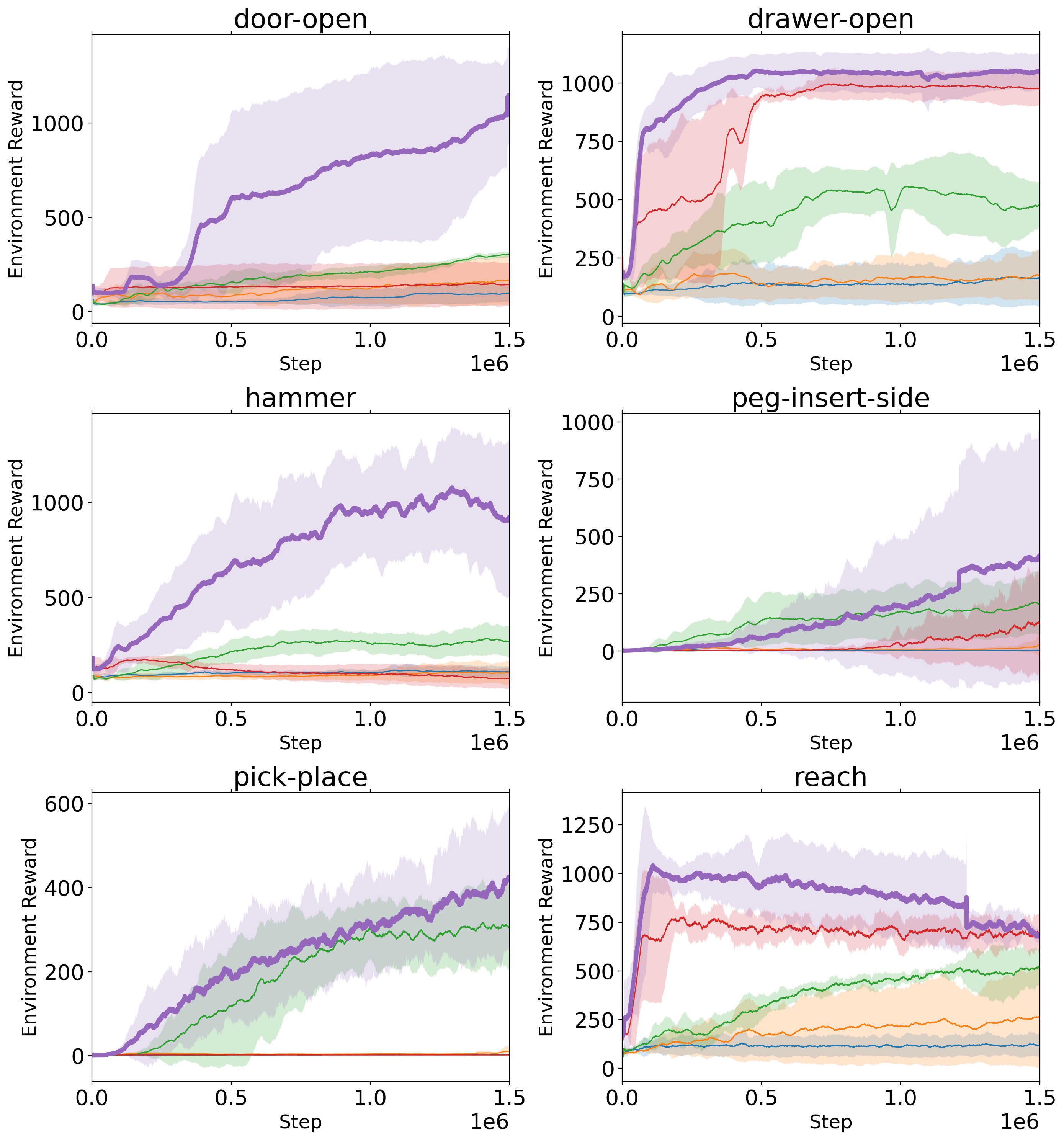}
    \caption{The evolution of the episodic reward} \label{fig:train-reward}
\end{subfigure}\hfill\vrule width 1pt\hfill
\begin{subfigure}{0.48\linewidth}
    \centering
    \includegraphics[width=1.0\linewidth]{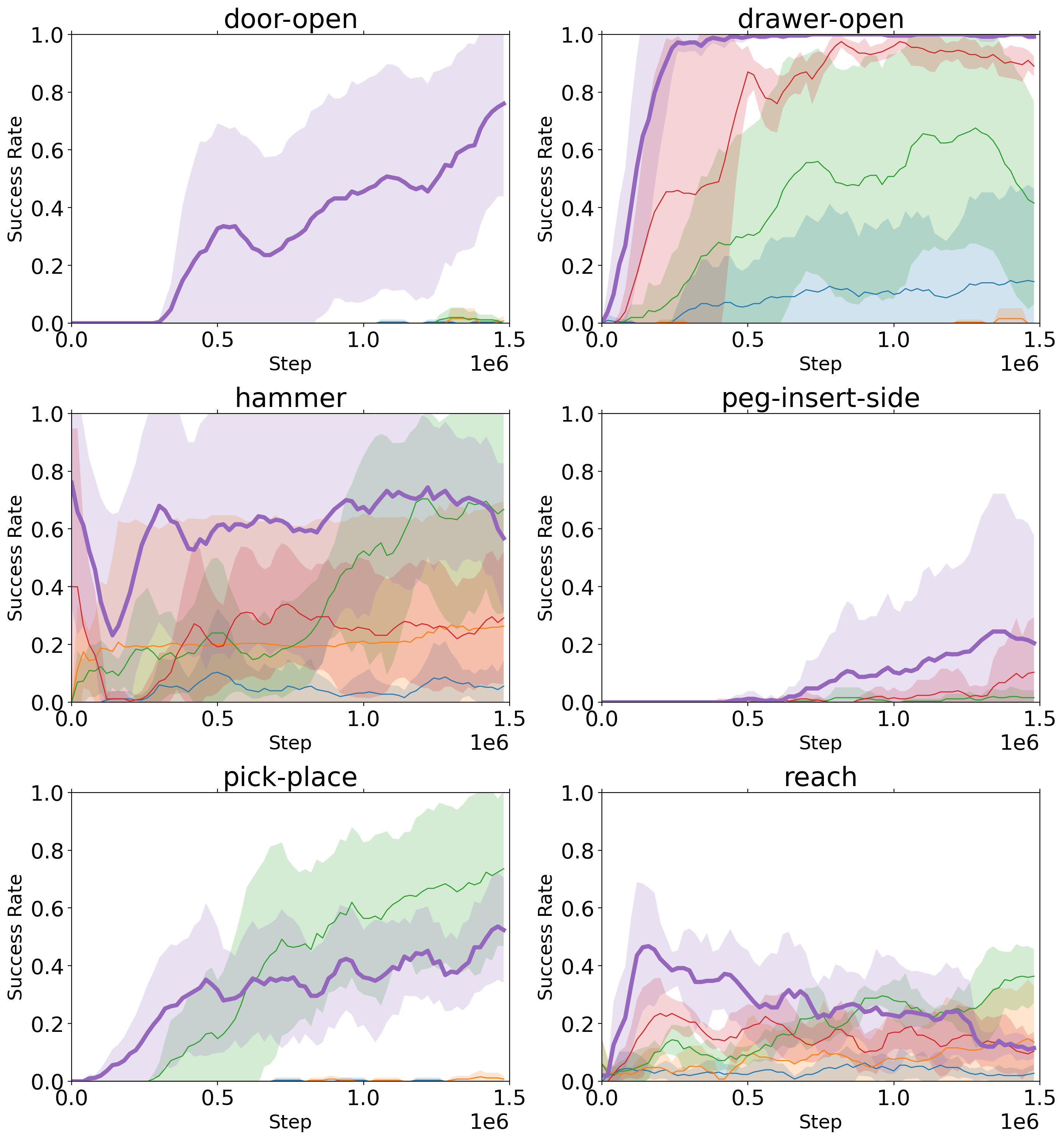}
    \caption{The evolution of the success rate 
    }
    \label{fig:success-rate}
    \end{subfigure}
\caption{Visualization of policy learning in the Meta-World \cite{yu2020meta} simulation environment. We run \alg and several baselines on six diverse tasks of various difficulties. We also run \alg without online push-back as an ablation. We report the environment reward during training (\textbf{left}) and the task success rate from 10 rollouts (\textbf{right}) averaged over five seeds. The solid line denotes the mean and the transparent area denotes standard deviation. \alg demonstrates clear advantages in both metrics, especially at early stages of training.}
\label{fig:meta_world_result}
\end{figure*}

In our simulated experiments, we used benchmark tasks from the Meta-World environment~\citep{yu2020meta}, selecting six table-top manipulation tasks (Figure~\ref{fig:all-tasks} e-j): \texttt{door-open}, \texttt{drawer-open}, \texttt{hammer}, \texttt{peg-insert-side}, \texttt{pick-place}, and \texttt{reach}. For all tasks, we trained the model-free reinforcement learning algorithm DrQ-v2~\citep{yarats2021image} using the reward predicted by \alg in place of the environment reward.%

We compare our method to three relevant baselines: (1) \textbf{TCN}~\cite{sermanet2018time}, a temporal contrastive method that optimizes embeddings for measuring task progress; (2) \textbf{GAIL}~\cite{ho2016generative}, an adversarial approach that aligns the state-action distribution of a policy with that of an expert. GAIL's discriminator is used as a reward function for our baseline following \citet{yang2024rank2reward}; 
and (3) \textbf{Rank2Reward}~\cite{yang2024rank2reward}, a state-of-the-art reward learning method that combines temporal ranking and classification of expert demonstrations to estimate rewards. For a fair comparison, we use their default hyperparameters and trained all methods under the same setup. All methods were trained for 1.5M steps, relying solely on the reward estimates provided by the model.

We evaluate the methods in terms of the episodic return, which reflects the cumulative reward received by the agent during each episode of the task. To ensure the robustness of our results, we average the episodic return over five different seeds for both the baseline methods and our approach, providing a comprehensive comparison of performance across all tested tasks. This setup allows us to analyze the impact of our proposed method in a controlled, simulated environment.

We present the results in Figure~\ref{fig:meta_world_result}. Our method significantly outperforms the baseline methods across most tasks, with a notable advantage in efficiency. In the \texttt{drawer-open} and \texttt{hammer} tasks, \alg requires only 10\% of the total training budget to achieve higher rewards than the baselines. In the \texttt{door-open} and \texttt{peg-insert-side} tasks, where almost all other methods fail entirely, our approach demonstrates strong performance and a higher success rate.

\textbf{Ablating Push-back} 
We examine the impact of adversarial online refinement via push-back in our framework. Across various tasks (see Figure~\ref{fig:meta_world_result}), we find that online push-back significantly enhances \alg's performance, particularly on challenging tasks such as \texttt{door-open} and \texttt{hammer}. These tasks demand more precise discrimination between expert behavior and online roll-outs to provide an informative reward signal. Notably, \alg without push-back still outperforms every baseline on the \texttt{drawer-open} and \texttt{reach} tasks. 

In the case of the \texttt{reach} task, we observe that \alg with push-back shows a decrease in return as training progresses, a trend that is less pronounced in the version without push-back. We hypothesize that this occurs because the model initially achieves a high success rate---reaching a peak with only 10\% of the total steps---leading the refinement process to overly penalize the agent, even when its behavior closely resembles the expert. 

Online training with non-expert data is crucial for robust performance. TCN, which is only trained on expert demonstrations without online updates, fails at every task due to the domain shift between pre-training and online RL.

\subsection{Real-World Robotic Experiments}
\begin{table}[!ht]
  \centering
  \begin{tabular}{l p{0.5\linewidth}}
    \toprule
    Task & Behavior of Failed Demonstrations\\
    \midrule
    \multirow{3}{*}{\texttt{Drawer-Open}} & The robot approaches the drawer but misses the handle by a few centimeters, failing to open it.\\
    \hline
    \multirow{2}{*}{\texttt{Drawer-Close}} &  The robot misses the drawer by a few centimeters, failing to close it.\\
    \hline
    \multirow{3}{*}{\texttt{Push-Block}} & The robot moves the block halfway toward the target cup but stops before reaching it.\\
    \hline
    \multirow{3}{*}{\texttt{Pick-Place-Cup}} & The robot lowers to pick up the cup but misses it by a few centimeters and remains in that position. \\
    \bottomrule
  \end{tabular}
  \caption{Failure modes in the collected suboptimal trajectories for the real-world robotic experiments. For each task we collected 20 failed demonstrations to go with the 20 expert demonstrations.}
  \label{tab:real-robot-failure-descriptions}
\end{table}

In this section, we demonstrate how \alg, pretrained on the egocentric EPIC-KITCHENS dataset~\cite{Damen2018EPICKITCHENS, Damen2022RESCALING}, can efficiently learn robotic tasks from a limited number of demonstrations, even when some are unsuccessful. Our approach enhances sample efficiency and robustness to noisy data in offline RL, making it more effective than traditional behavior cloning (BC) methods.

\subsubsection{Pretraining on Kitchen Dataset}
Using ResNet34~\cite{he2016deep} as a backbone, we first pretrain our encoder $\mE_{\theta}$ with Equation~\ref{eqn:objective-function} taking P01-P07 sequences from the EPIC-KITCHENS dataset composed of approximately 1.29M frames. We randomly sample frame triplets $(\vo_i, \vo_j, \vo_g)$ from the videos ensuring a maximal frame gap $\lVert i-g \rVert \leq 2000$. To improve the robustness of $\mE_{\theta}$, we additionally train with distractor frames where we replace the current observation $\vo_j$ with a frame randomly sampled from different video sequence $\vo_{j'}$ as negative examples. For negative triplets, we replace $\mu$ in Eqn.~\ref{eqn:e_output} with $-1$ as the label. We train our model with batch size 128 for 30000 iterations using the Adam optimizer with constant learning rate $2e-4$. 

\subsubsection{Baselines}
We compare \alg with R3M~\cite{nair2022r3m} and VIP~\cite{ma2022vip}. VIP and R3M are self-supervised visual representations shown to provide dense reward functions for robotic tasks. Both R3M and VIP are pretrained on Ego4D~\cite{grauman2022ego4d} with 4.3M frames from videos from 72000 clips. R3M is trained via time contrastive learning, where the distance between frames closer in time is smaller than for images farther in time. Additionally, they leverage L1 weight regularization and language embedding consistency losses. VIP uses a contrastive approach treating frames close in time as negative pairs and those further away as negative pairs towards learning visual goal-conditioned
value functions.

\begin{figure}[t]
    \centering
    \includegraphics[width=\linewidth]{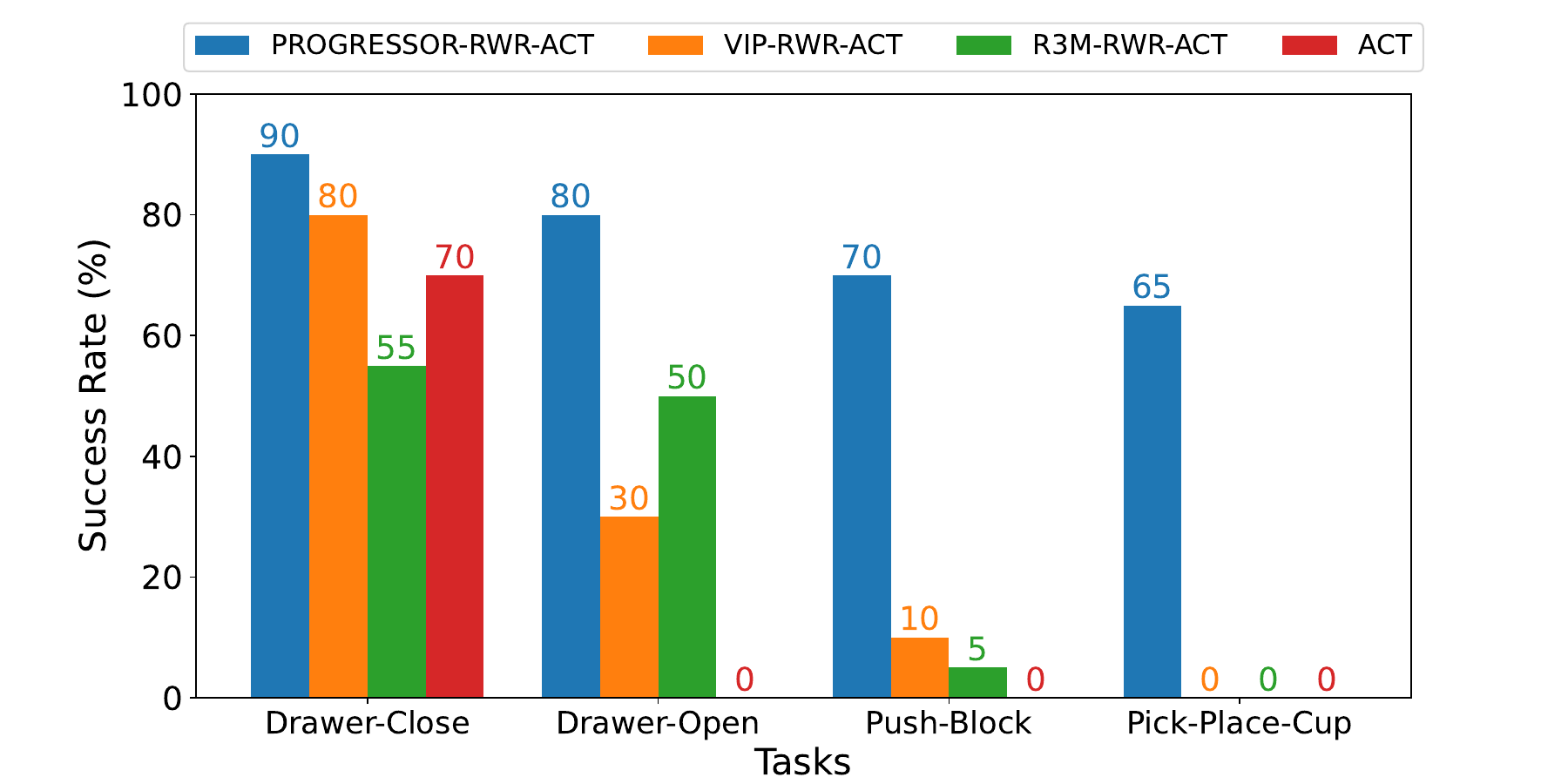}
    \caption{
    Success rates for four real-world tasks, where RWR-ACT is trained on a combination of correct and failed demonstrations using \alg, R3M, and VIP as reward models.
    } \label{fig:real-world-eval}
\end{figure}

\begin{figure*}[htbp]
    \centering
    \begin{subfigure}[b]{0.5\textwidth}
        \centering
        \includegraphics[width=\textwidth]{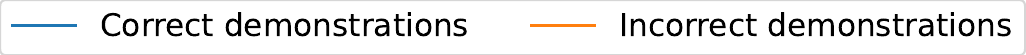}
    \end{subfigure}
    
    \begin{subfigure}[b]{0.331\textwidth}
        \centering
        \includegraphics[width=\textwidth]{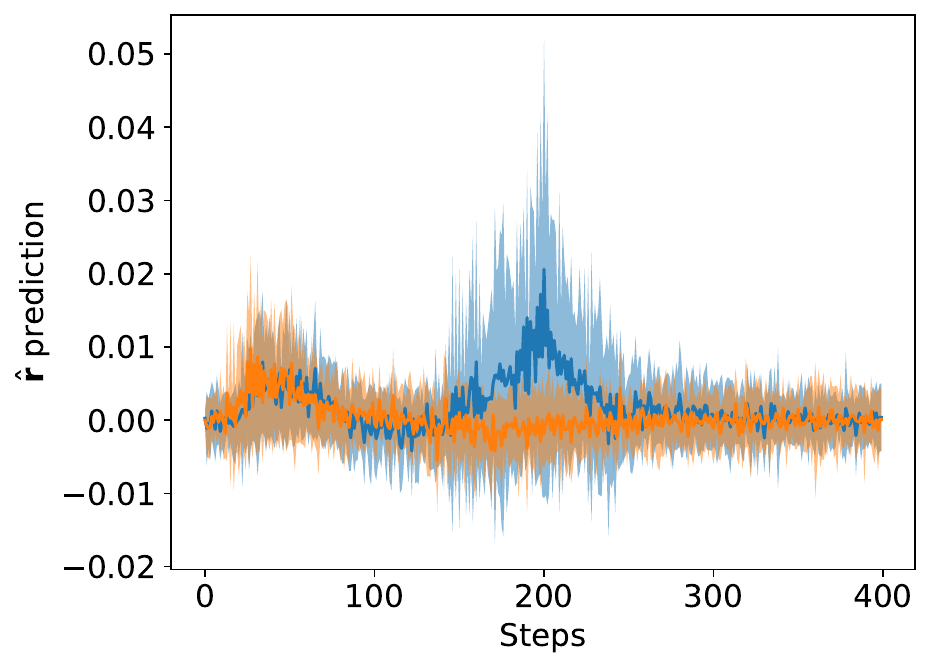}
        \caption{R3M}
    \end{subfigure}%
    \hfill
    \begin{subfigure}[b]{0.315\textwidth}
        \centering
        \includegraphics[width=\textwidth]{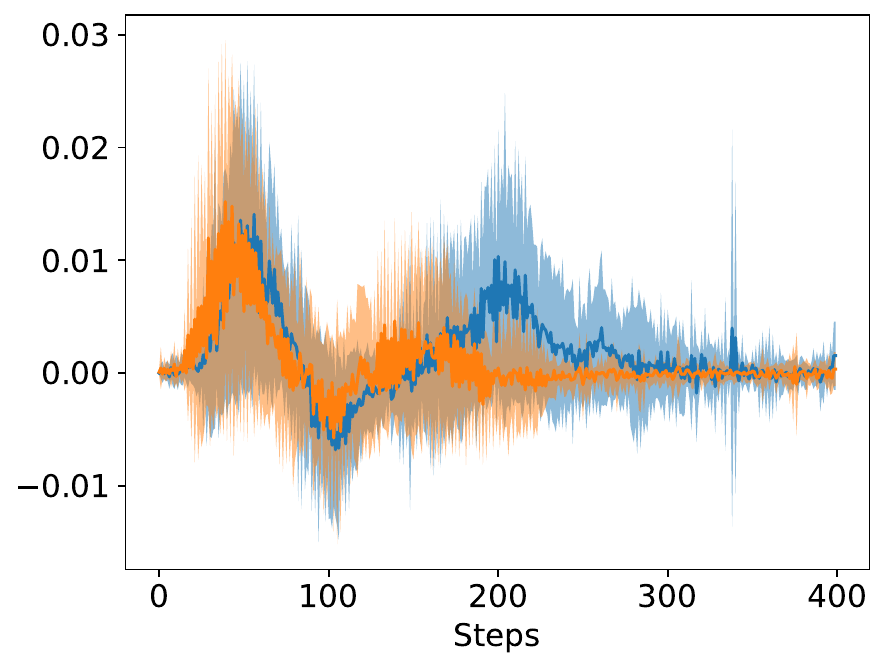}
        \caption{VIP}
    \end{subfigure}%
    \hfill
    \begin{subfigure}[b]{0.299\textwidth}
        \centering
        \includegraphics[width=\textwidth]{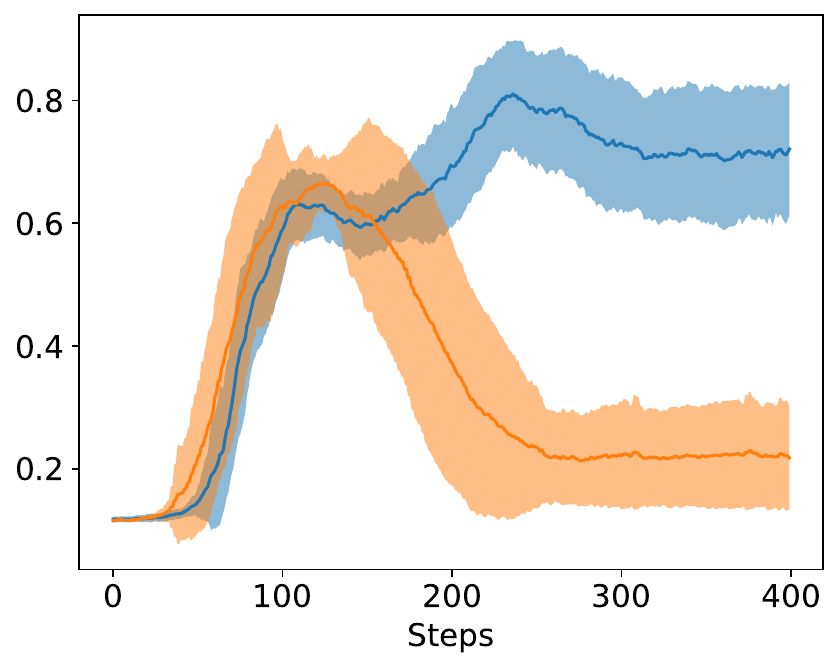}
        \caption{\alg}
    \end{subfigure}
    \caption{Mean reward predictions, $\hat{\vr}$, for \textcolor{myblue}{correct} and \textcolor{myorange}{incorrect} demonstrations in the \texttt{Drawer-Open} task from \alg, R3M, and VIP. \alg provides more distinct reward weighting between correct and failed trajectories compared to the baseline models.}
    \label{fig:failed-correct-reward}
\end{figure*}

\subsubsection{Real-World Few-Shot Offline Reinforcement Learning with Noisy Demonstrations}
Following the offline reinforcement learning experiments of \citet{ma2022vip}, we leveraged the reward-weighted regression (RWR) method \cite{peters2007reinforcement, peng2019advantage}. Our aim in applying RWR is to show that a reward model trained on human videos can help robots learn to perform tasks even when the training data contains noisy, suboptimal trajectories. In scenarios with highly suboptimal trajectories, vanilla behavior lloning (BC) approaches often struggle to learn the correct behaviors. However, goal-conditioned reward weighting provides a signal that focuses learning from the correct sub-trajectories, enabling the visual imitation learning model to effectively learn accurate action executions.

To this end, we start with a vanilla BC model and adapt the loss function to be weighted by a pretrained reward model. Specifically, we employed Action-Chunking Transformer (ACT)~\citep{zhao2023learning}, a BC model designed for learning fine-grained manipulation tasks. We modified ACT’s reconstruction loss—originally defined as the mean absolute error between the predicted action and the ground-truth action—to incorporate RWR as follows:
\begin{equation} \label{eqn:rwr}
   \mathcal{L}_{\text{reconst}} = \exp\left(\omega \cdot \hat{\mathbf{r}}\right) \cdot 
   \left\| \pi_{\text{ACT}}(\vo_j) - a_t \right\|_{1},
\end{equation}
where $\hat{\mathbf{r}}$ is the reward prediction for the current observation $\vo_{j}$, \(\omega\) is the temperature parameter, \(a_t\) is the ground-truth action, and \(\pi_{\text{ACT}}(\vo_j)\) represents the action predicted by ACT by taking the current observation image ($\vo_j$) as an input. In this paper, we refer to ACT trained with the reconstruction loss replaced by Equation~\ref{eqn:rwr} as RWR-ACT, to distinguish it from the standard ACT.

We compare \alg with R3M and VIP by freezing the pre-trained models and using them as reward prediction models to train RWR-ACT on downstream robotic learning tasks. The reward predictions from these models are used in place of $\mathbf{u}_{\theta}$ in Equation~\ref{eqn:rwr}. In both VIP and R3M, the reward prediction is parameterized by the current observation, previous observation, and goal observation. In contrast, our method parameterizes reward using the initial observation, current observation, and goal observation. 
Additionally, we include vanilla ACT as a baseline, which applies uniform weighting (i.e., $\omega=0$) across all observations in the training trajectories .

We design four tabletop robotic manipulation tasks (see Figure~\ref{fig:all-tasks} a-d): \texttt{Drawer-Open}, \texttt{Drawer-Close}, \texttt{Pick-Place-Cup}, and \texttt{Push-Block}. %
For each task, we collect 40 demonstrations, half of which are suboptimal and fail to complete the task. Including these failed demonstrations is crucial for evaluating whether the learned reward model can accurately signal progress toward a goal by assigning high reward to transitions that lead toward completion and low reward to those that do not. 
Table~\ref{tab:real-robot-failure-descriptions} summarizes the behaviors in failed demonstrations. Detailed task and data descriptions along with frame sequences from both successful and failed trajectories, are provided in S9.
We categorize \texttt{Drawer-Close} as an easy task and the other three as hard tasks. This distinction in difficulty is based on the complexity of the tasks and the level of suboptimality in failed demonstrations.

We train all policies using the same hyperparameters employed by \citet{zhao2023learning} for training ACT in their real-world behavior cloning experiments (full details in S9.2).
For all RWR experiments, we set $\omega=0.1$. Each method is then evaluated over 20 rollouts, and the success rate is reported. The success criteria for each task are presented in S9.

The evaluation results presented in Figure~\ref{fig:real-world-eval} highlight the significant advantage of \alg-RWR-ACT, which consistently outperforms all baseline methods across all tasks. 
This advantage is particularly pronounced in the more challenging tasks, such as \texttt{Push-Block} and \texttt{Pick-Place-Cup}, on which other methods struggle to achieve meaningful task performance. While R3M-RWR-ACT and VIP-RWR-ACT show moderate success in simpler tasks, such as \texttt{Drawer-Close}, their performance significantly drops in \texttt{Push-Block} and \texttt{Pick-Place-Cup} tasks. As can be seen in Figure~\ref{fig:failed-correct-reward}, the gap in reward prediction between failed and correct trajectories for \alg is notably larger compared to R3M and VIP, starting when the robot begins to deviate from the goal observation (around step 125). This supports our hypothesis that \alg provides more distinct reward weighting between both correct and failed trajectories compared to the baseline models. 
Vanilla ACT performs poorly, especially in difficult tasks. Since ACT assigns equal weight to all sub-trajectories, it cannot prioritize more relevant transitions or filter out noise effectively, leading to poor task execution.

\begin{figure}[t]
    \centering
    \begin{subfigure}{\linewidth}
    \includegraphics[width=0.97\linewidth]{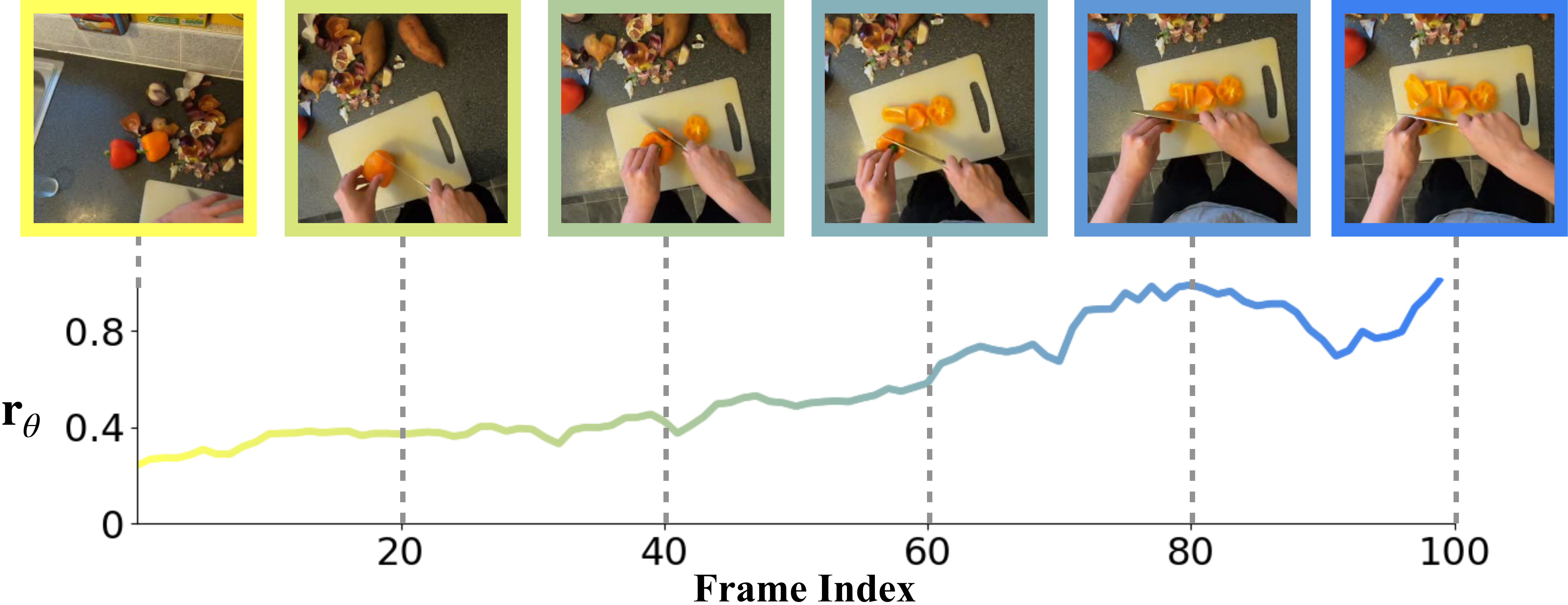}
    \caption{The evolution of the episodic reward on human video} \label{fig:train-reward-human}
    \end{subfigure}
    
    \begin{subfigure}{\linewidth}
    \vspace{1em}
    \includegraphics[width=0.97\linewidth]{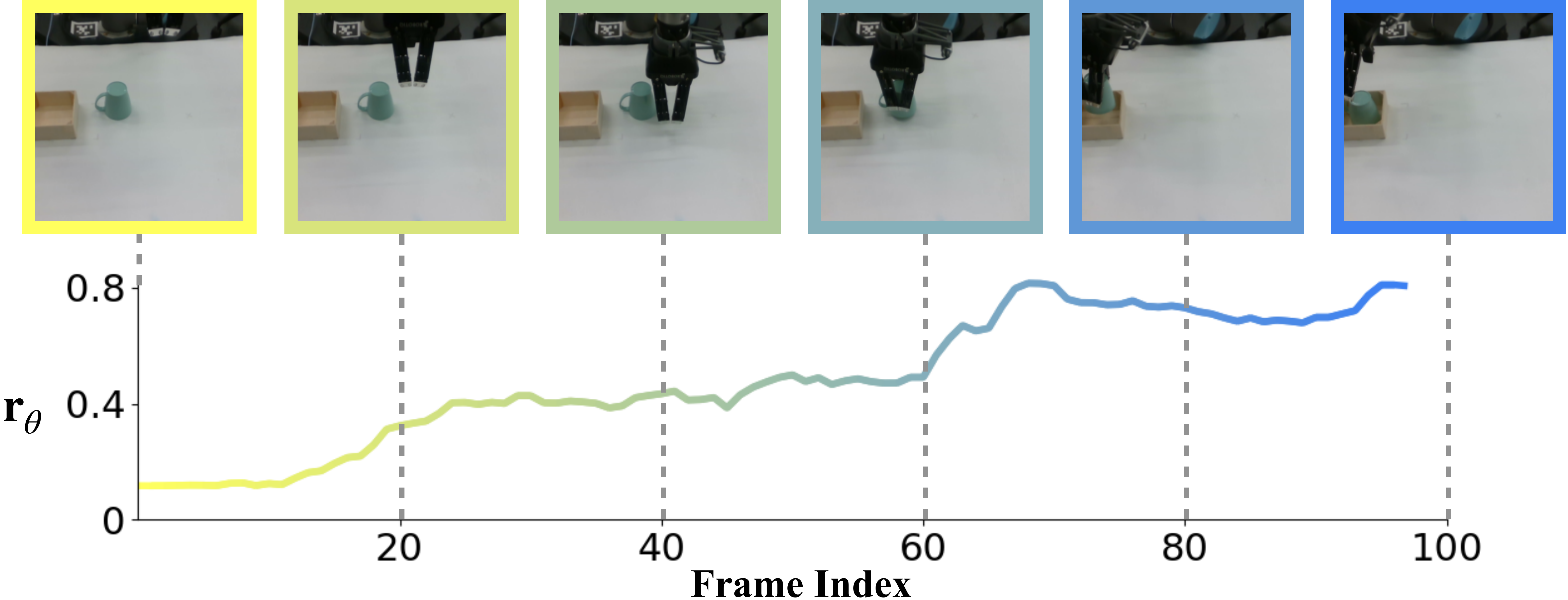}
    \caption{The evolution of the episodic reward on robot video} \label{fig:train-reward-robot}
    \end{subfigure}
    \caption{Visualization of the predicted rewards from Epic Kitchen pretrained \alg, evaluated on a sample clip from EPIC-KITCHENS (\textbf{top}) and on sample frames from an unseen \texttt{Pick-Place-Cup} demonstration on our robot (\textbf{bottom}).}
    \label{fig:progress-plot}
\end{figure}

Figure~\ref{fig:progress-plot} shows reward predictions for a sample frame sequence from EPIC-KITCHENS and a trajectory from our \texttt{Pick-Place-Cup} robotic task. The figure shows that, despite being pretrained solely on human videos, \alg generates well-formed reward predictions for a robotic task in a zero-shot setting.

\section{Conclusion and Limitations}\label{sec:conclusion}
In this work, we presented \alg, a self-supervised framework that learns task-agnostic reward functions from video via progress estimation. By learning progress estimation between observations in expert trajectories, \alg generates dense rewards that effectively guide policy learning. During online RL training, this progress estimation is further refined through an adversarial push-back strategy, helping the model handle non-expert observations and minimize distribution shift. Our method shows enhanced performance compared to previous state-of-the-art approaches across a range of simulated experiments.
In real-robot experiments, we applied \alg, pretrained on in-the-wild human videos, to learn policies with limited and noisy task demonstrations, outperforming other visual reward models. 

\textbf{Limitations and Future Work}
We acknowledge several limitations in our work. (1) Our method assumes a linear progression of tasks and models progress as a unimodal prediction, making it unsuitable for tasks with cyclic observations, such as those in the DeepMind Control Suite~\cite{tassa2018deepmind}. Future enhancements that model progress as a multimodal prediction could address this limitation more effectively. (2) While our online RL experiments demonstrate that our hyperparameters are robust across various tasks, incorporating a dynamic weighting factor \(\beta\) for refinement may further enhance performance.

\clearpage
\clearpage
\bibliographystyle{ieeenat_fullname}
\bibliography{main}

\clearpage
\setcounter{page}{1}
\maketitlesupplementary

\section{\alg Training Details}
\subsection{Architecture and Training}
\label{sup:progressor-training-details}
\alg can, in principle, be trained with any visual encoding architecture, requiring only minor modifications to the final layer to predict Gaussian parameters. In our experiments, we utilize the standard ResNet34 model~\cite{he2016deep}, replacing its final fully-connected layer with an MLP of size [$512$, $512$, $128$]. Given that our method processes triplets of inputs $(\vo_i, \vo_j, \vo_g)$, the resulting representation has a size of [$128 \times 3$]. This representation is then fed into an MLP with layers of size [$128 \times 3$, 2048, 256]. Finally, two prediction heads are derived from the 256-dimensional output, predicting $\mu$ and $\log\sigma^2$.

We pretrain \alg for $30000$ steps using the EPIC-KITCHENS dataset~\cite{Damen2018EPICKITCHENS} for the real-world experiments, and for $10000$ steps for  experiments performed in simulation. In the pretrianing steps for both our simulated and real-world experiments, we first sample a trajectory (i.e., a video clip) from the pretraining dataset. We then randomly select an initial frame $\vo_i$ as well as a goal frame $\vo_g$ from the selected trajectory such that $\lVert g -i \rVert \leq 2000$. Finally, we uniformly randomly select an intermediate frame $\vo_j$, where $i < j < g$.%

\subsection{Hyperparameters}
\begin{table}[!h]
    \centering
    \begin{tabular}{ccc}
        \toprule
        & Simulation & Real-World RWR\\
        \midrule
        $\alpha$ & $0.4$ & $0.4$\\
        $\beta$ & $0.9$ & $-$\\     
        \bottomrule
    \end{tabular}
    \caption{Hyperparameters used by \alg for experiments performed in simulation and the real world.}
    \label{tab:hyperparameters}
\end{table}
Table~\ref{tab:hyperparameters} lists the hyperparameters used for both the simulation online RL and real-world offline RL experiments. A consistent $\alpha = 0.4$ is used across all experiments. The push-back decay factor is employed exclusively in the simulated online RL experiments.

\section{Simulation Experiment Details}
\label{sup:simultation-experiment-details}

In this section, we describe the tasks and the data generation process employed using the MetaWorld environment~\cite{yu2020meta} for our simulation experiments.

\subsection{Meta-World Tasks}
We took six diverse tasks from the Meta-World environment~\cite{yu2020meta}, described in Table~\ref{tab:meta-world-tasks}. In all tasks, the position of the target object, such as the drawer or hammer, is randomized between episodes.

\begin{table}[!ht]
  \centering
  {\footnotesize
  \begin{tabular}{l l}
    \toprule
    Task & Task Description \\
    \midrule
    \texttt{door-open} & Open a door with a revolving joint. \\
    \texttt{drawer-open} & Open a drawer.  \\
    \texttt{hammer} & Hammer a screw on the wall. \\
    \texttt{peg-insert-side} & Insert a peg sideways. \\
    \texttt{pick-place} & Pick and place a puck to a goal. \\
    \texttt{reach} & Reach a goal position. \\
    \bottomrule
  \end{tabular}}
  \caption{Meta-World \cite{yu2020meta} task descriptions.}
  \label{tab:meta-world-tasks}
\end{table}

\subsection{Expert Data Generation}
To collect expert trajectories for our simulated experiments, we execute Meta-World's oracle policies. For each task, we generated 100 successful rollouts for training and 10 for testing.  This dataset is subsequently used for pretraining \alg in our simulated experiments, following the steps outlined in Section~\ref{sup:progressor-training-details}. 

\section{Real-World Robot Experiment Details}
\label{sup:real-robot-experiment-details}

\subsection{Robotic Experiment Setup}
\begin{figure}[!t]
    \centering
    \includegraphics[width=0.9\linewidth]{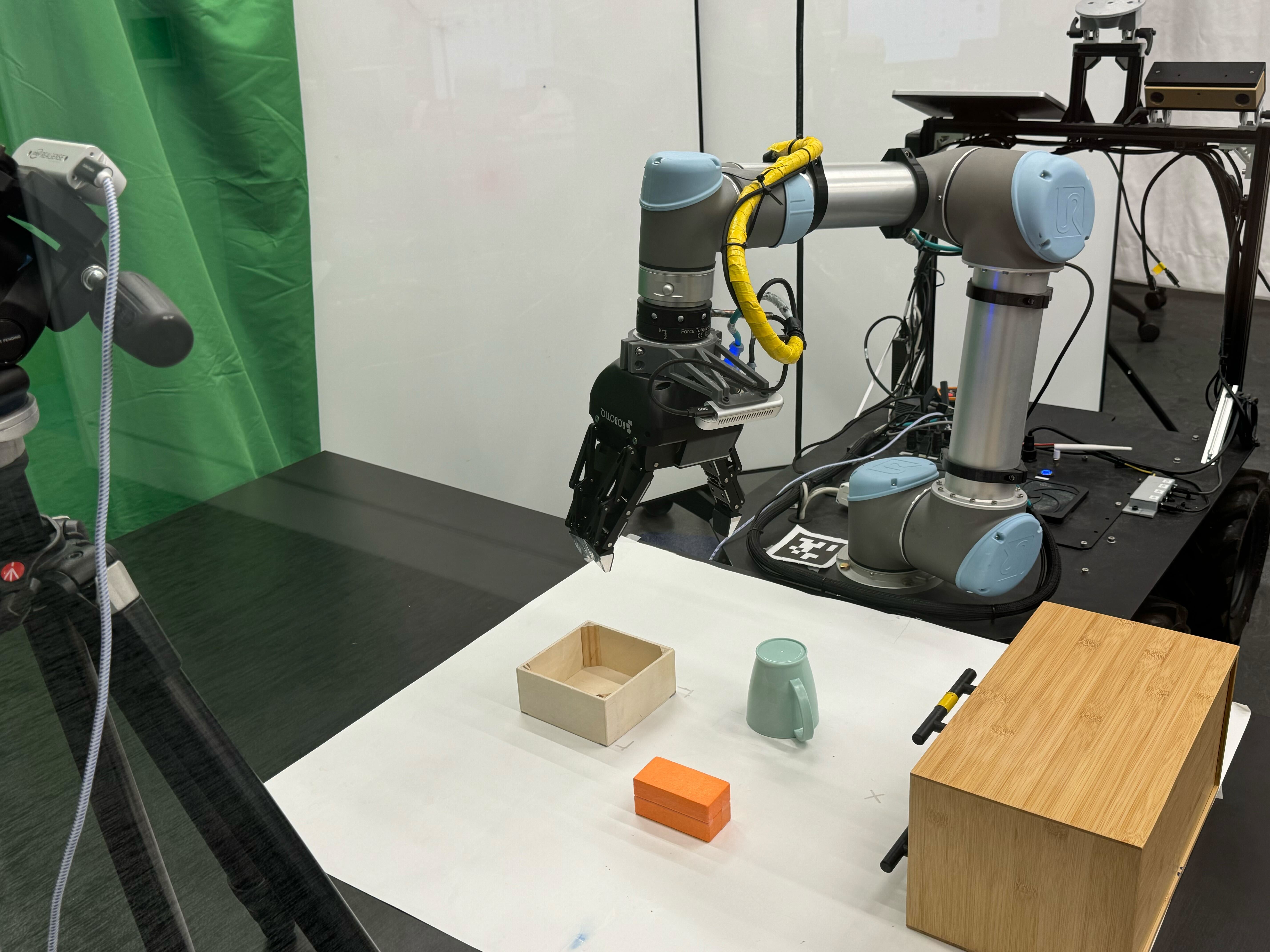}
    \caption{The real-world experiments were conducted using a Universal Robots UR5 robot arm equipped with a Robotiq~3-Finger Gripper. One RealSense camera is mounted to the end-effector and the other RealSense camere is fixed in the environment using a tripod (visible in the left of the image). } \label{fig:robot-setup}
\end{figure}
The real-robot experiments are performed using a Universal Robots UR5 robot arm equipped with a Robotiq 3-Finger Gripper (Figure~\ref{fig:robot-setup}). The setup includes two RealSense cameras: one mounted on the robot's wrist that images the gripper, and the second on a fixed tripod facing the robot.

\subsection{Robotic Demonstration Data Collection}
\begin{figure*}[!ht]
    \centering
    \begin{subfigure}[b]{\textwidth}
        \centering
        \include{figs/tasks/drawer_open}
    \end{subfigure}
    \begin{subfigure}[b]{\textwidth}
        \centering
        \include{figs/tasks/drawer_close}
    \end{subfigure}
    \begin{subfigure}[b]{\textwidth}
        \centering
        \include{figs/tasks/push_block}
    \end{subfigure}
    \begin{subfigure}[b]{\textwidth}
        \centering
        \include{figs/tasks/pick_place_cup}
    \end{subfigure}
    \caption{\textcolor{mygreen}{Correct} and \textcolor{myred}{incorrect} demonstrations (every 80th frame) for each real-robot task from a third-person camera view used in our experiments.
    To see how \alg and the baselines differentiate between correct and incorrect trajectories, see Figure \ref{fig:failed-correct-reward} (\texttt{Drawer-Open}) and Figure \ref{fig:failed-correct-reward-supp} (other three tasks).} \label{fig:correct-incorrect-demonstrations}
\end{figure*}

We collect demonstrations by teleoperating the UR5 using a Meta Quest~3 controller~\cite{q2r}. We record each demonstration at 30\,Hz for $400$ 
steps. Figure~\ref{fig:correct-incorrect-demonstrations} shows video frame sequences from correct and incorrect demonstrations for the four real-world tasks.

\subsection{Task Descriptions}
\begin{figure}[!t]
    \centering
    \begin{subfigure}[b]{0.22\textwidth}
        \centering
        \includegraphics[width=\textwidth]{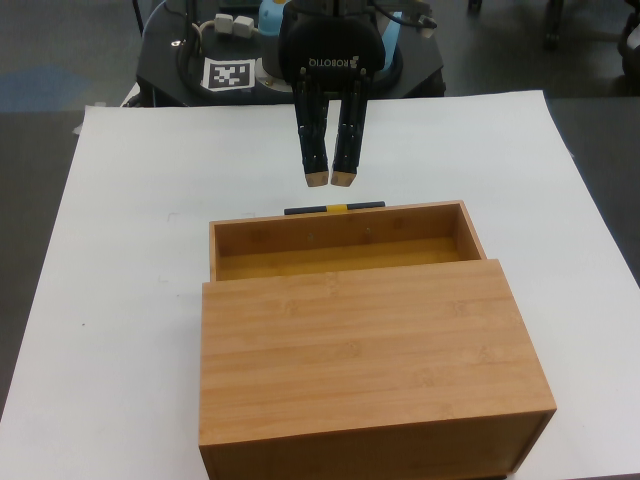} 
        \caption{\texttt{Drawer-Open}}
    \end{subfigure}
    \hfill
    \begin{subfigure}[b]{0.22\textwidth}
        \centering
        \includegraphics[width=\textwidth]{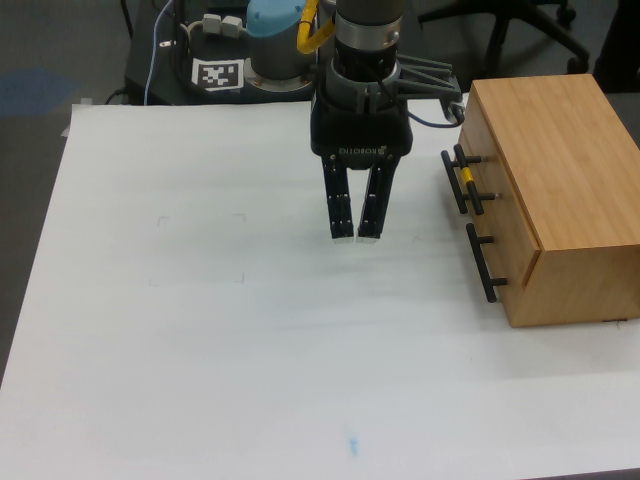}
        \caption{\texttt{Drawer-Close}}
    \end{subfigure}
    \hfill
    \begin{subfigure}[b]{0.22\textwidth}
        \centering
        \includegraphics[width=\textwidth]{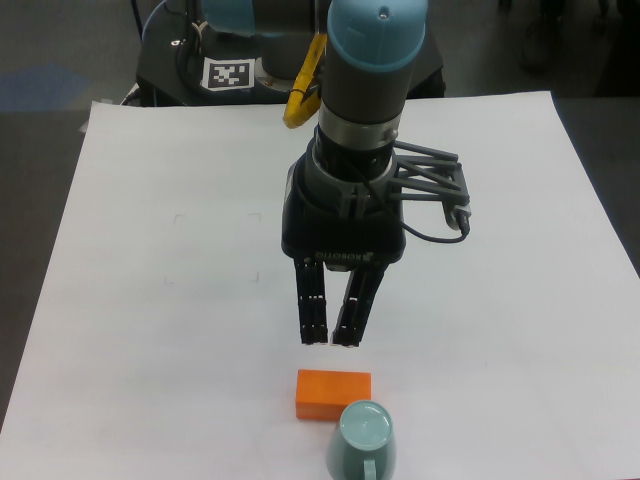}
        \caption{\texttt{Push-Block}}
    \end{subfigure}
    \hfill
    \begin{subfigure}[b]{0.22\textwidth}
        \centering
        \includegraphics[width=\textwidth]{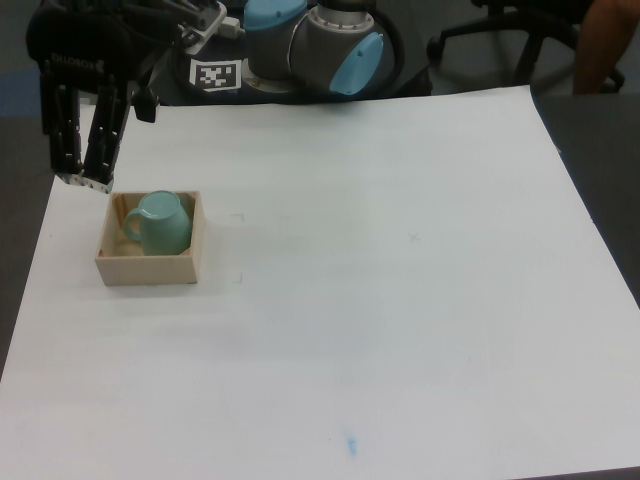} 
        \caption{\texttt{Pick-Place-Cup}}
    \end{subfigure}
    \caption{The goal images that we use for each task for reward weighting in the offline reinforcement learning experiments.}
    \label{fig:goal_frames}
\end{figure}

The tasks involve a variety of object manipulation challenges designed to test reward weighting in offline Reinforcement Learning. In the \texttt{Drawer-Close} task, the objective is to close a drawer starting from an open position. The \texttt{Drawer-Open} task requires the agent to pull a drawer open from a closed state. In the \texttt{Push-Block} task, the goal is to push a block toward a specified target, a cup, ensuring the block moves into proximity with the cup. Finally, \texttt{Pick-Place-Cup} involves picking up a cup and carefully placing it into a designated box. Figure~\ref{fig:goal_frames} displays the goal frames representing the completion of each task.

\subsection{Training and Evaluation Details}
\begin{table}[!ht]
  \centering
  {\footnotesize
  \begin{tabular}{l l}
    \toprule
    Task & Success Criterion \\
    \midrule
    \texttt{Drawer-Open} & Drawer is open by more than $5$\,cm. \\
    \texttt{Drawer-Close} & Drawer is within $1$\,cm of being fully closed. \\
    \texttt{Push-Block} & Block is within $5$\,cm of the cup. \\
    \texttt{Pick-Place-Cup} & Cup is placed inside the box. \\
    \bottomrule
  \end{tabular}}
  \caption{Success criterion for the real-robot experiments.}
  \label{tab:success-criterion}
\end{table}
Our few-shot offline RL implementation builds upon the Action-Chunking Transformer (ACT)~\cite{zhao2023learning}. The inputs to the model include (i) two 640$\times$480 RGB images from the RealSense cameras, and (ii) proprioceptive data consisting of the 6-DoF joint angles and the binary open or close state of the gripper. The action space consists of the 6-DoF translational and rotational velocity of the end-effector and a binary open or close command of the gripper.

In the reward-weighted regression (RWR) setup, rewards are computed by providing all reward models with the final frame of a correct demonstration from each task as the goal image. For all reward predictions, frames from the fixed RealSense camera were used. Figure~\ref{fig:goal_frames} illustrates the goal images used for the four robotic tasks, which were consistently used across all reward predictions. The temperature scale in RWR was set to \(\omega = 0.1\) for all tasks.

\begin{table}[H]  %
  \centering
  \begin{tabular}{lc}
    \toprule
    Hyperparameter & Value \\
    \midrule
    prediction horizon & 30 \\
    learning rate & $10^{-5}$ \\
    batch size & 64 \\
    epochs & 5000 \\
    $\omega$ & 0.1\\   
    \bottomrule
  \end{tabular}
  \caption{Hyperparameters for reward-weighted ACT training.}
  \label{tab:real-robot-hyperparams}
\end{table}

Table~\ref{tab:real-robot-hyperparams} shows the hyperparameters used to train the \{\alg, VIP, R3M\}-RWR-ACT models. For the vanilla ACT, we used the same hyperparameters as listed in Table~\ref{tab:real-robot-hyperparams}, except for $\omega = 0$.

During inference, the model predicts a sequence of actions (a ``chunk'') of length 30 and then executes each action before predicting the next chunk. We did not use a temporal ensemble as proposed by \citet{zhao2023learning}, since we found that it causes the gripper to drift and negatively impacts performance.

For evaluation, we conduct 20 test rollouts for each task and report the success rate. The success criteria for each task are outlined in Table~\ref{tab:success-criterion}.

\section{Ablation}
\begin{figure}[!ht]
    \centering
    \includegraphics[width=\linewidth]{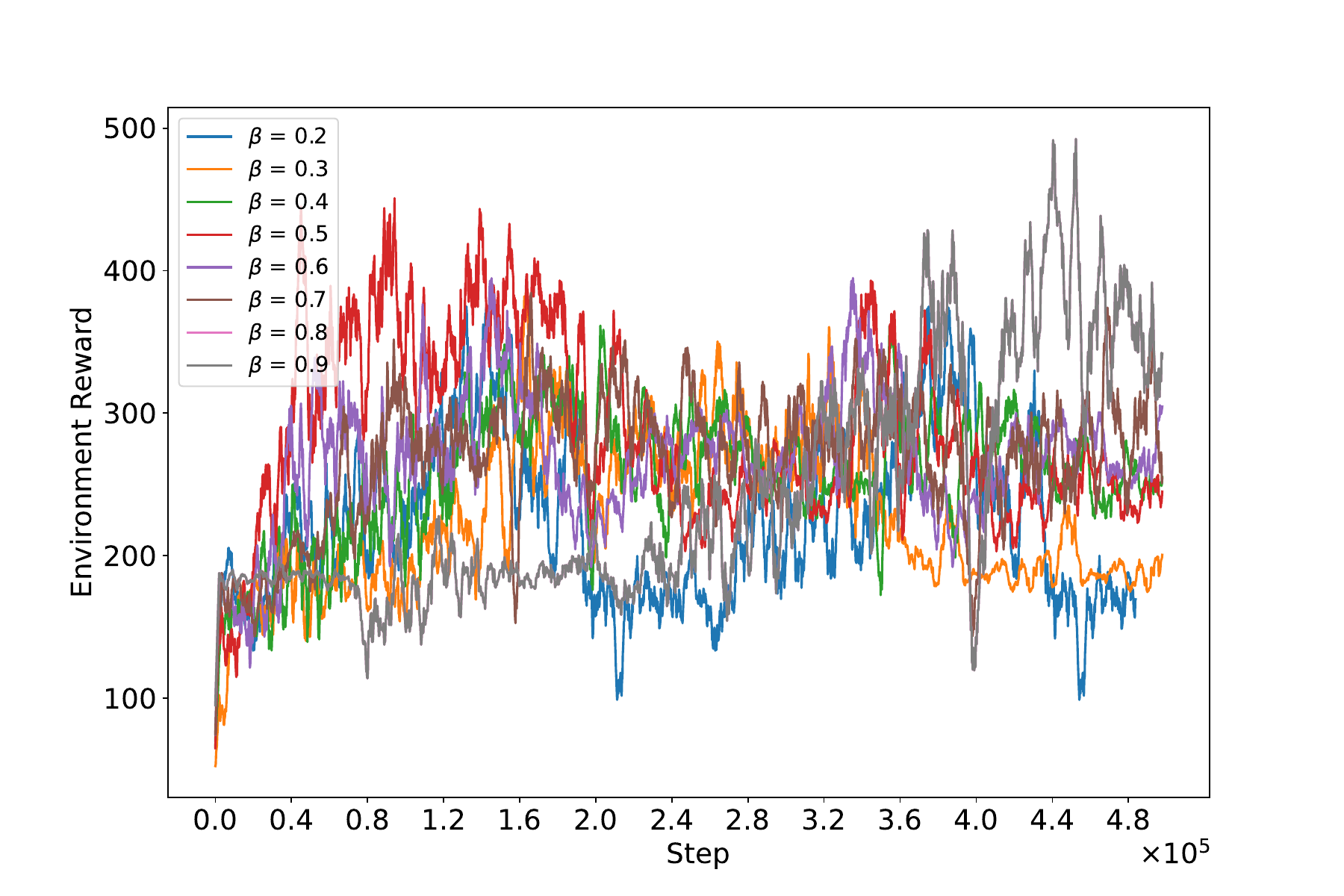}
    \caption{Visualization of policy learning performance in Meta-World's \texttt{hammer} environment, evaluating the effect of different values for the push-back decay factor $\beta$. The plot highlights the accumulated rewards over training, demonstrating how varying $\beta$ values influences the agent's ability to optimize performance.
    }
    \label{fig:push_back_ablation}
\end{figure}

In this section, we present an ablation study evaluating different values of the push-back decay factor ($\beta$) while training a DrQ-v2 agent on Meta-World's \texttt{hammer} task, using a fixed seed of 121. The case of $\beta=0$ (\alg without Push-back) is discussed in the main paper. Figure~\ref{fig:push_back_ablation} depicts the environment rewards accumulated during training. As shown in the figure, the agent achieves higher rewards with $\beta = 0.9$.

\section{Qualitative Analysis}
\begin{figure}[!t]
    \centering
    \begin{subfigure}{\linewidth}
        \centering
        \includegraphics[width=\linewidth]{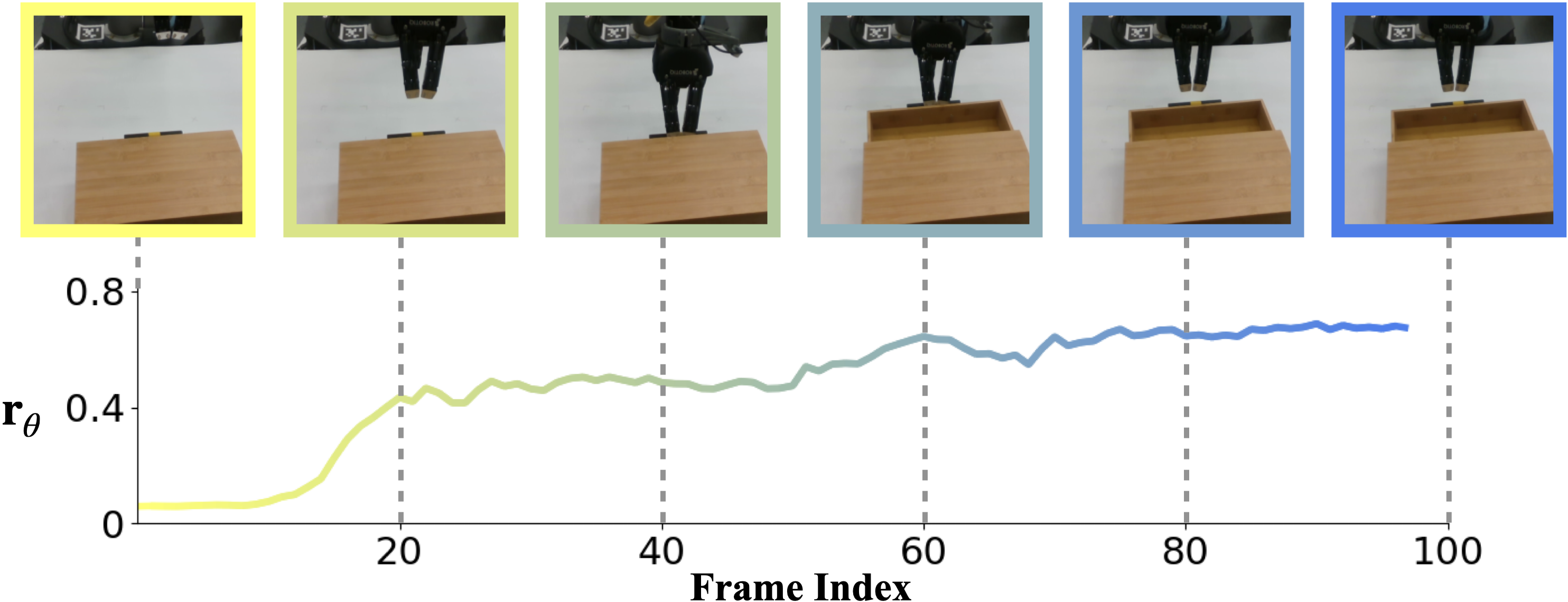}
        \caption{Reward plot for \texttt{Drawer-Open} task}
    \end{subfigure}
    \vfill
    \vspace{2em}
    \begin{subfigure}{\linewidth}
        \centering
        \includegraphics[width=\linewidth]{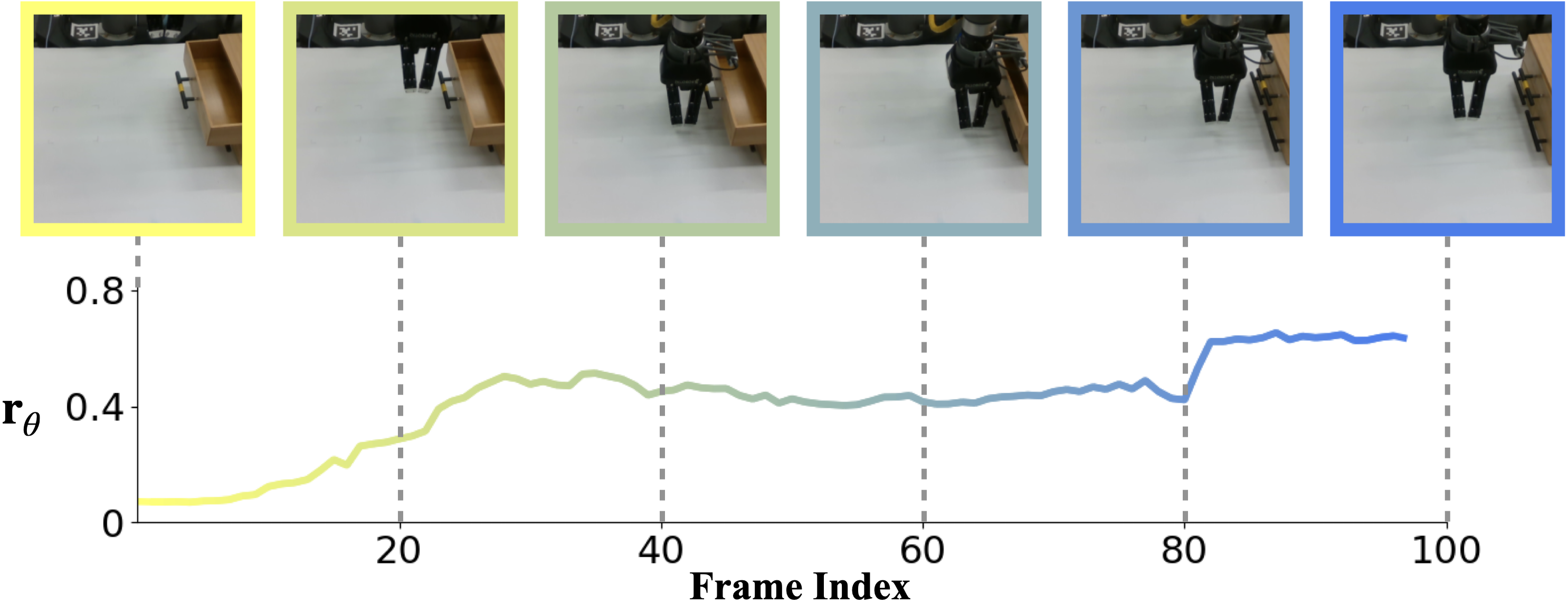}
        \caption{Reward plot for \texttt{Drawer-Close} task}
    \end{subfigure}
    \vfill
    \vspace{2em}
    \begin{subfigure}{\linewidth}
        \centering
        \includegraphics[width=\linewidth]{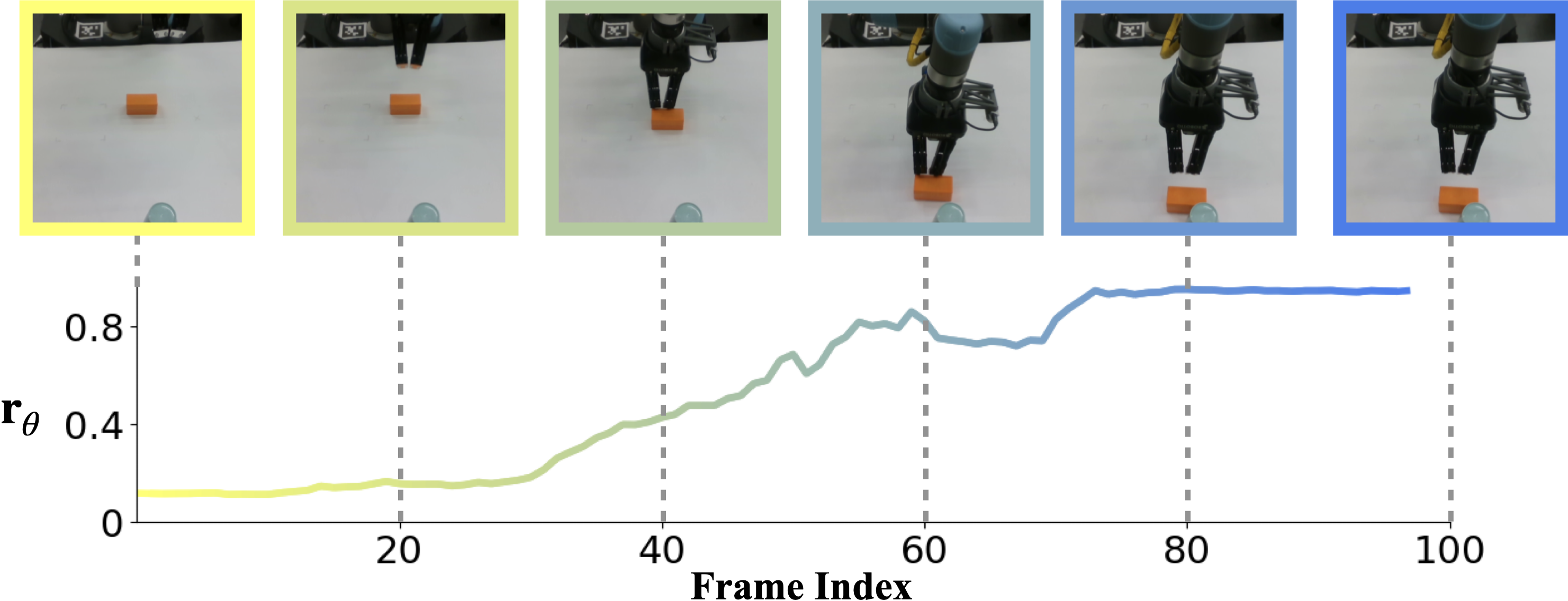}
        \caption{Reward plot for \texttt{Push-Block} task}
    \end{subfigure}
    \caption{Visualization of the predicted reward by \alg, pretrained on EPIC-KITCHENS and evaluated zero-shot on correct robotic demonstrations for (\textbf{a}) \texttt{Drawer-Open} (\textbf{b}) \texttt{Drawer-Close} and (\textbf{c}) \texttt{Push-Block}   tasks.}
    \label{fig:progress-plot-sup}
\end{figure}

Figure~\ref{fig:progress-plot-sup} presents zero-shot reward predictions from \alg pretrained on the EPIC-KITCHENS dataset. This figure serves as an extension to Figure~\ref{fig:progress-plot} for completeness. It includes zero-shot reward predictions for sample correct trajectories from our collected real-robot demonstrations for the tasks \texttt{Drawer-Open}, \texttt{Drawer-Close}, and \texttt{Push-Block}.
\begin{figure*}[htbp]
    \centering
    \begin{subfigure}[b]{0.5\textwidth}
        \centering
        \includegraphics[width=\textwidth]{figs/legend.pdf}
    \end{subfigure}
    
    \begin{subfigure}[b]{0.331\textwidth}
        \centering
        \includegraphics[width=\textwidth]{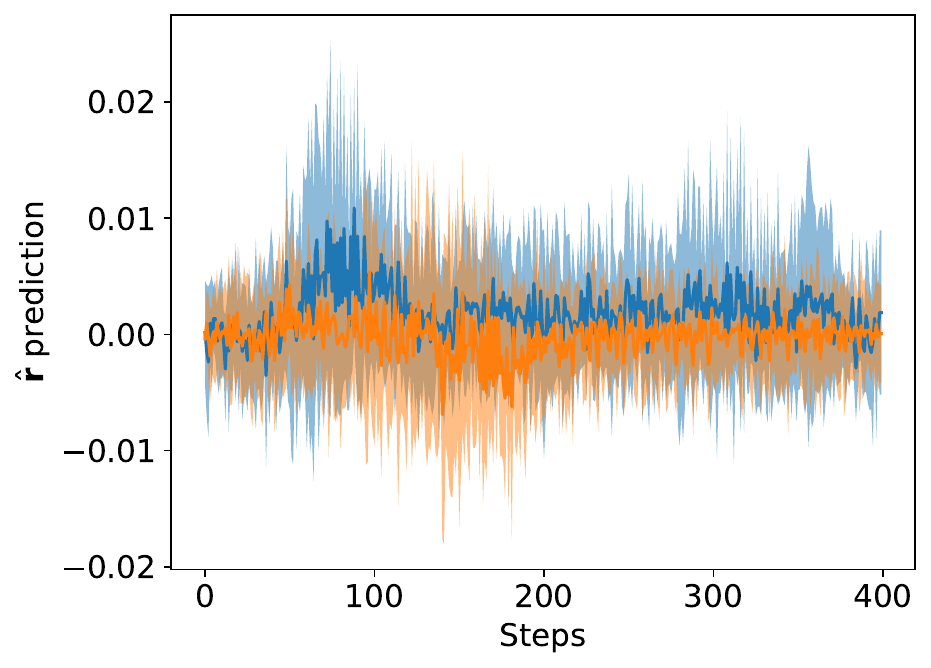}
        \caption{R3M - \texttt{Drawer-Close} }
    \end{subfigure}%
    \hfill
    \begin{subfigure}[b]{0.315\textwidth}
        \centering
        \includegraphics[width=\textwidth]{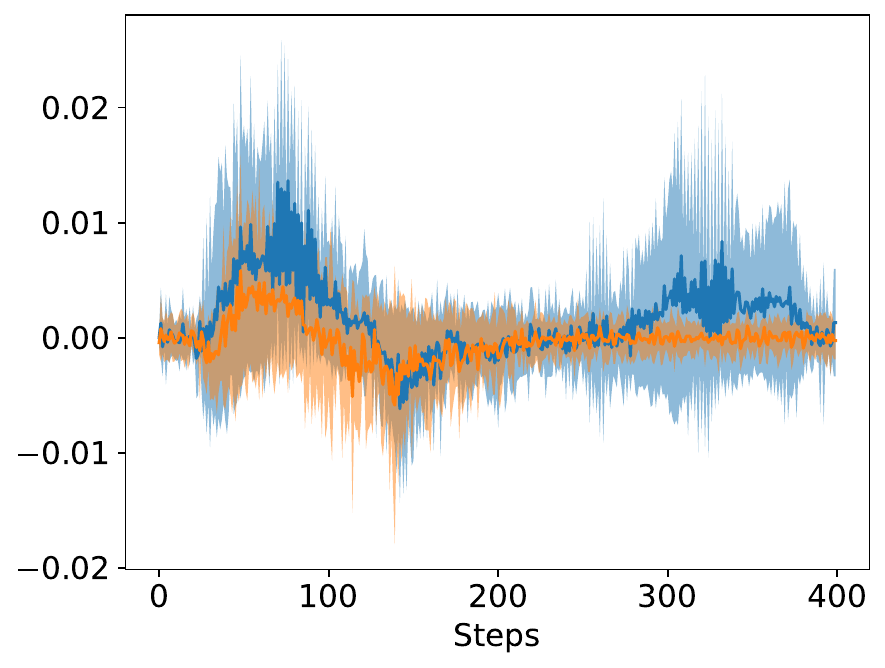}
        \caption{VIP - \texttt{Drawer-Close} }
    \end{subfigure}%
    \hfill
    \begin{subfigure}[b]{0.299\textwidth}
        \centering
        \includegraphics[width=\textwidth]{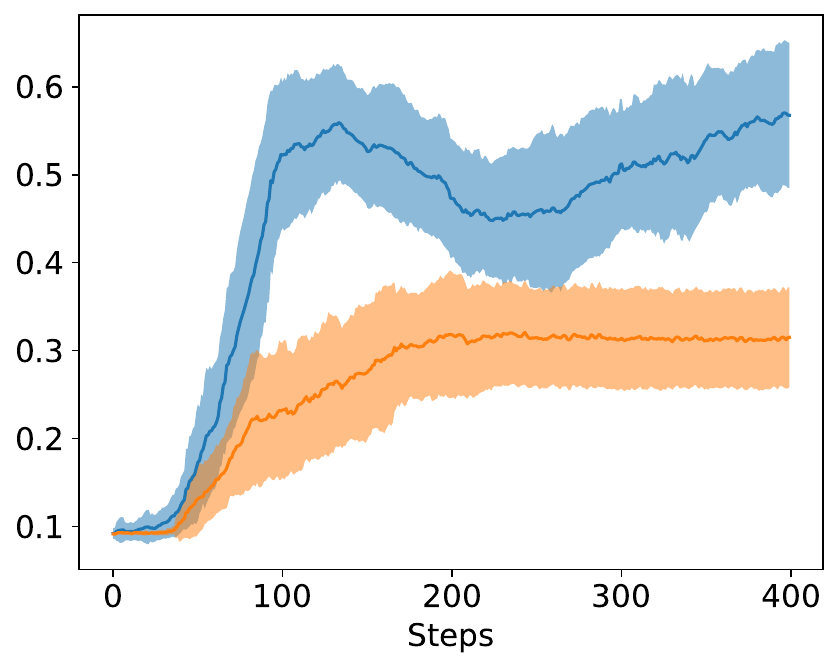}
        \caption{\alg - \texttt{Drawer-Close} }
    \end{subfigure}

    \vspace{1em} %
    \begin{subfigure}[b]{0.331\textwidth}
        \centering
        \includegraphics[width=\textwidth]{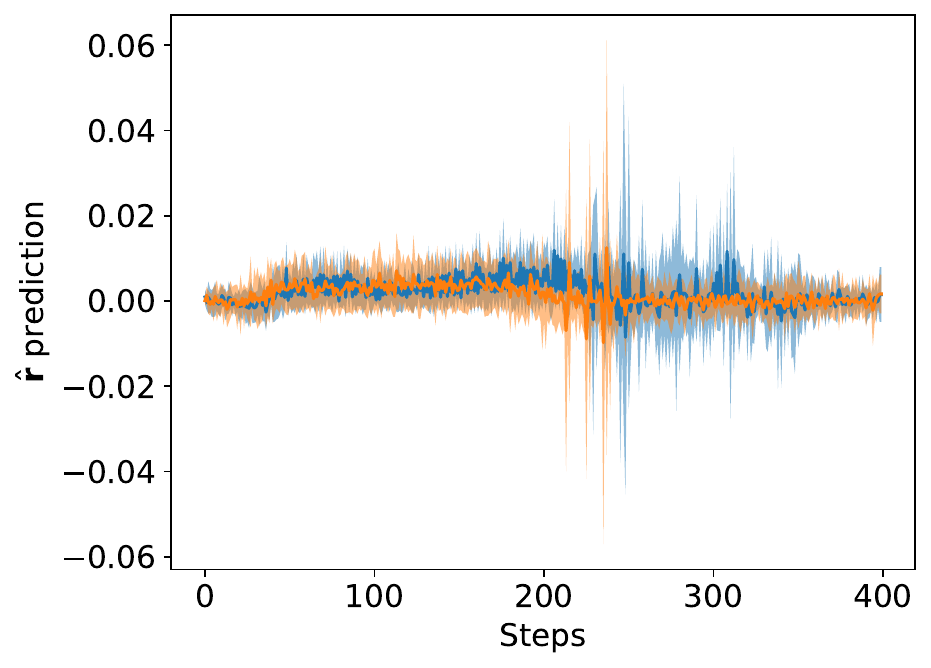}
        \caption{R3M - \texttt{Push-Block} }
    \end{subfigure}%
    \hfill
    \begin{subfigure}[b]{0.315\textwidth}
        \centering
        \includegraphics[width=\textwidth]{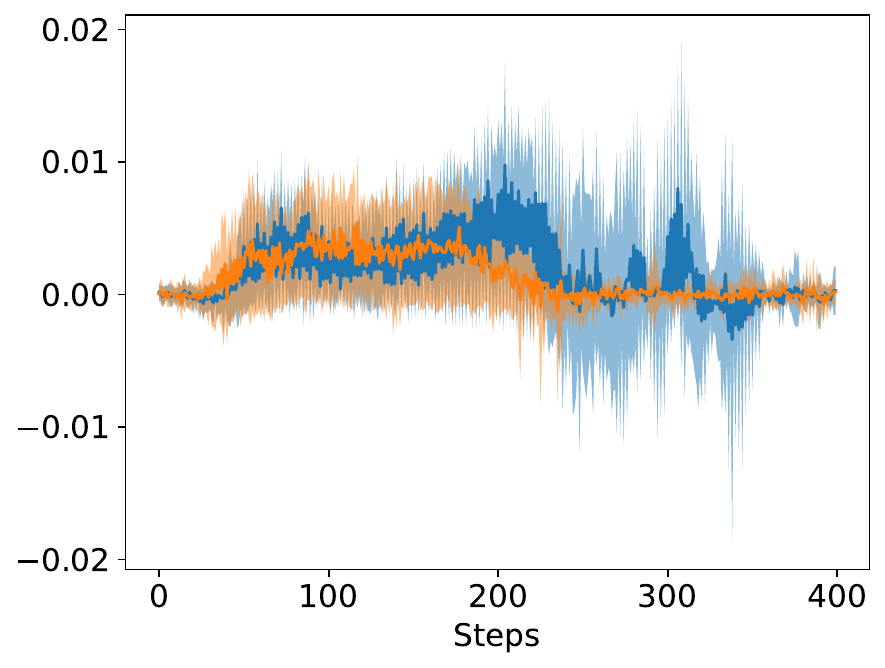}
        \caption{VIP - \texttt{Push-Block} }
    \end{subfigure}%
    \hfill
    \begin{subfigure}[b]{0.299\textwidth}
        \centering
        \includegraphics[width=\textwidth]{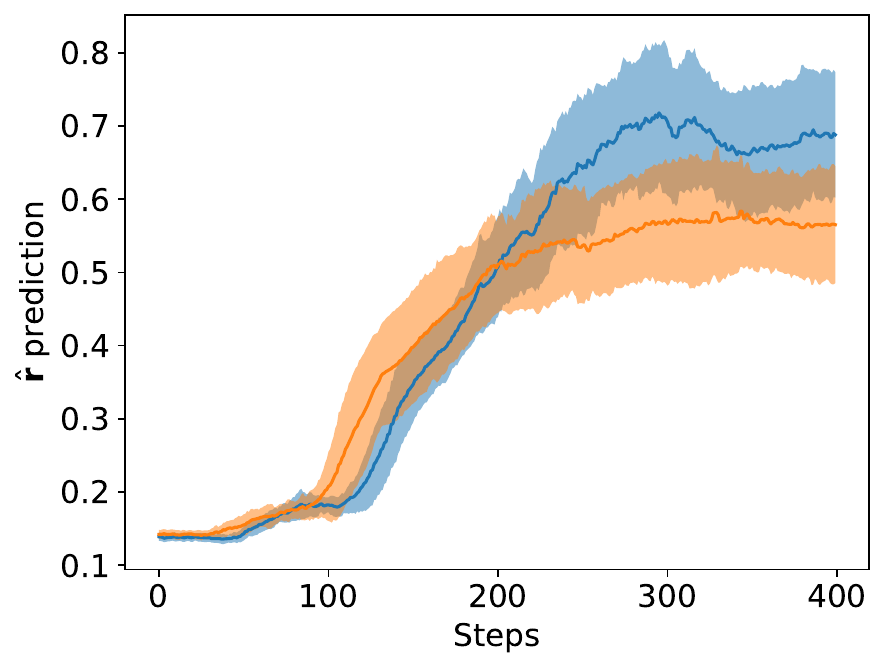}
        \caption{\alg - \texttt{Push-Block} }
    \end{subfigure}

    \vspace{1em} %
    \begin{subfigure}[b]{0.331\textwidth}
        \centering
        \includegraphics[width=\textwidth]{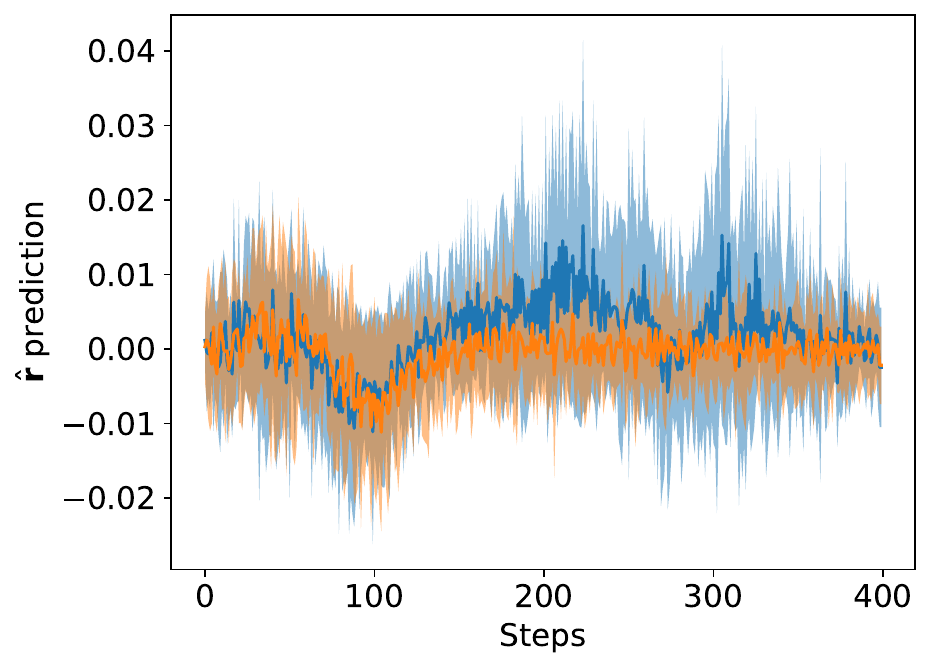}
        \caption{R3M - \texttt{Pick-Place-Cup} }
    \end{subfigure}%
    \hfill
    \begin{subfigure}[b]{0.315\textwidth}
        \centering
        \includegraphics[width=\textwidth]{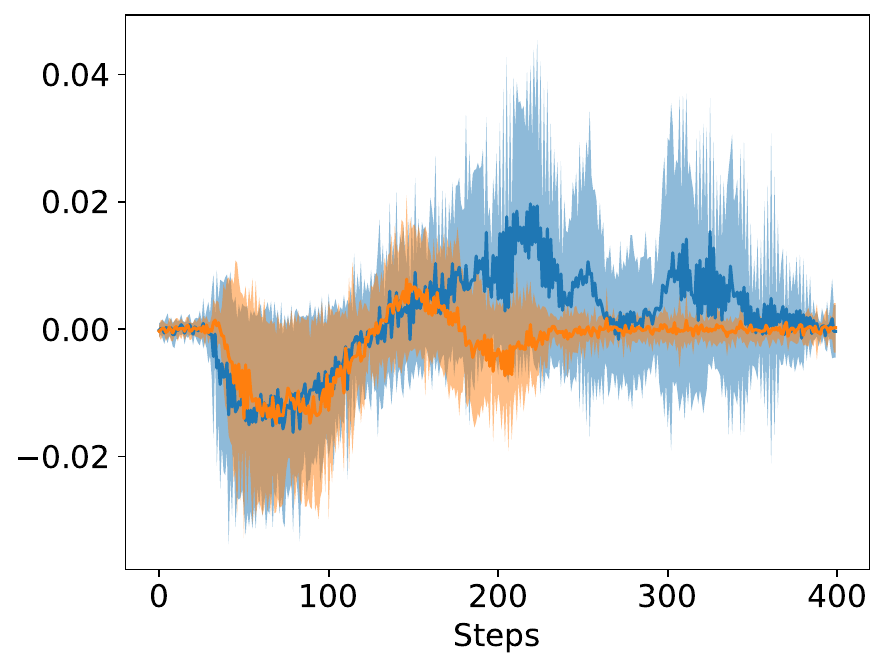}
        \caption{VIP - \texttt{Pick-Place-Cup} }
    \end{subfigure}%
    \hfill
    \begin{subfigure}[b]{0.299\textwidth}
        \centering
        \includegraphics[width=\textwidth]{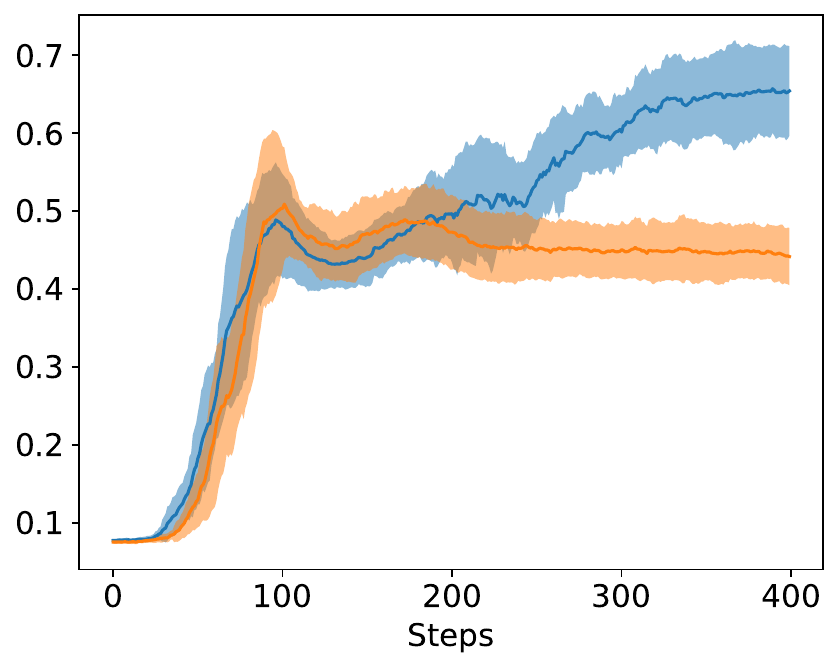}
        \caption{\alg - \texttt{Pick-Place-Cup} }
    \end{subfigure}

    \caption{Mean reward predictions $\hat{\vr}$ of (\textbf{left column}) R3M, (\textbf{middle column})  VIP, and (\textbf{right column}) \alg for \textcolor{myblue}{correct} and \textcolor{myorange}{incorrect} demonstrations for the \texttt{Drawer-Close}, \texttt{Push-Block}, and \texttt{Pick-Place-Cup} tasks. \alg provides reward predictions (weights) that better differentiate between  correct and incorrect trajectories, consistently outperforming the baseline models.}
    \label{fig:failed-correct-reward-supp}
\end{figure*}

For completeness, we have included plots similar to the mean reward prediction figure from the main paper (see Figure~\ref{fig:progress-plot}). The plots in Figure~\ref{fig:failed-correct-reward-supp} provide a qualitative comparison of reward predictions from our robotic demonstration dataset for three additional real-world tasks: \texttt{Drawer-Close}, \texttt{Push-Block}, and \texttt{Pick-Place-Cup}. As illustrated in the figure, for all tasks, our reward model consistently predicts lower average rewards for the sub-trajectories in the incorrect demonstrations, where the failures occur.

\end{document}